%% file: neurips_2026.tex
\documentclass{article}

\usepackage[main, final]{neurips_2026}

\usepackage{enumitem}
\usepackage[utf8]{inputenc} 
\usepackage[T1]{fontenc}    
\usepackage{hyperref}       
\usepackage{url}            
\usepackage{booktabs}       
\usepackage{wrapfig}
\usepackage{amsfonts}       
\usepackage{nicefrac}       
\usepackage{microtype}      
\usepackage{xcolor}         

\usepackage{graphicx}
\usepackage{subcaption}
\usepackage{amsmath}
\usepackage{amssymb}
\usepackage{mathtools}
\usepackage{amsthm}
\usepackage{multirow}
\usepackage[capitalize,noabbrev]{cleveref}
\usepackage[textsize=tiny]{todonotes}

\theoremstyle{plain}

\theoremstyle{definition}

\theoremstyle{remark}

\title{AdaST: Adaptive Coupling for Spatial-Temporal Forecasting}

\author{%
  Zhenyu Lei \\
  University of Virginia \\
  Charlottesville, VA, USA \\
  \texttt{vjd5zr@virginia.edu} \\
  \And
  Chenghao Liu\thanks{This work was completed prior to joining Datadog.} \\
  Datadog \\
  Paris, France \\
  \texttt{twinsken@gmail.com} \\
  \And
  Yushun Dong \\
  Florida State University \\
  Tallahassee, FL, USA \\
  \texttt{yd24f@fsu.edu} \\
  \And
  Qi R. Wang \\
  Northeastern University \\
  Boston, MA, USA \\
  \texttt{q.wang@northeastern.edu} \\
  \And
  Jundong Li \\
  University of Virginia \\
  Charlottesville, VA, USA \\
  \texttt{jundong@virginia.edu} \\
}

\begin{document}

\maketitle
\begin{abstract}
Spatial-temporal (ST) forecasting underpins many real-world systems such as traffic, climate, and energy networks. While existing methods implicitly assume strong spatiotemporal coupling, we observe that real-world ST data exhibits distinct coupling regimes, ranging from temporal-dominated and spatial-dominated to strongly coupled patterns.
This mismatch causes current models to suffer from spurious dependencies and degraded performance when one correlation dominates. To overcome this limitation, we aim to dynamically modulate spatial and temporal modeling based on the data's inherent coupling structure. However, three key challenges exist: unknown coupling structure, heterogeneous coupling dynamics, and suboptimal spatial modeling. We propose AdaST, an adaptive ST forecasting framework that tackles these challenges through a decompose-recompose paradigm.
AdaST factorizes inputs into components capturing different coupling patterns using heterogeneity-aware experts. Each component is processed by role-aligned modules, and a correlation-informed adaptive recomposer integrates them for final prediction. Extensive experiments confirm that AdaST significantly outperforms state-of-the-art baselines, validating the necessity of an adaptive approach.
\end{abstract}

\section{Introduction}
Spatial-temporal data, which incorporates both spatial and temporal information, plays a critical role in numerous real-world applications~\cite{wang2020deep, atluri2018spatio, birant2007st, li2025typhoformer}. Forecasting based on such data has proven invaluable across diverse domains by leveraging the inherent correlations among data points distributed across space and time~\cite{li2021spatial, lei2025st, guo2021learning}.
Spatial-temporal data is characterized by two fundamental correlations: spatial correlation, describing interactions among different locations, and temporal correlation, capturing how historical observations influence future states~\cite{vuran2004spatio}. These correlations are often intertwined, giving rise to complex spatial-temporal dependencies that complicate forecasting~\cite{shao2022decoupled, yi2024deep}.
To capture such dependencies, existing research has proposed various spatiotemporal coupling frameworks. Representative approaches include graph neural networks~\cite{zhao2026graphip} combined with temporal convolutional networks~\cite{diao2019dynamic}, attention-based architectures~\cite{zhang2019spatial}, and diffusion-based frameworks~\cite{yang2024survey}. Despite architectural differences, these approaches share a common design philosophy: spatial and temporal correlations are jointly modeled at each layer, implicitly assuming strong and homogeneous coupling throughout the data.

However, real-world ST data exhibits substantially different coupling structures across domains and scenarios~\cite{he2017space}.
Through systematic analysis, we categorize spatial-temporal data into three regimes based on coupling strength. \textbf{(1)} In weakly-coupled temporal-dominated scenarios, historical patterns at each location independently govern future values with minimal cross-location influence. For example, household energy consumption depends primarily on its own historical habits rather than neighboring behaviors~\cite{pierce2010home}.
\textbf{(2)} In weakly-coupled spatial-dominated scenarios, neighboring locations in short time windows exert primary influence while historical trends become less relevant. For example, traffic incidents immediately trigger downstream congestion regardless of historical patterns~\cite{qi2018dynamic}.
\textbf{(3)} In strongly-coupled scenarios, both correlations are tightly intertwined and predictions require balanced integration of temporal evolution and spatial diffusion. For example, urban air quality prediction requires balancing local historical emissions and wind-driven diffusion from neighbors~\cite{wang2018deep}.
Uniformly applying coupled frameworks across these structures leads to systematic performance degradation. For temporal-dominated data, forced spatial coupling introduces spurious cross-location dependencies. For spatial-dominated data, enforced temporal coupling incorporates misleading historical correlations. Even for strongly-coupled data, fixed architectures cannot adapt to heterogeneous coupling strengths across regions or time periods. Our preliminary experiments in Section~\ref{sec:prelim} further validate that mismatches between coupling structures and model assumptions significantly impair forecasting accuracy.

These observations motivate a fundamental question: Can we develop adaptive forecasting frameworks that dynamically modulate spatial and temporal modeling according to the data's inherent coupling structure? Nevertheless, addressing this question is challenging for three reasons. \textbf{(1) Unknown Coupling Structure.} The intrinsic coupling structure of a dataset is typically unknown a priori, where practitioners must rely on trial-and-error or domain expertise to select suitable architectures. \textbf{(2) Heterogeneous Coupling Dynamics.} Coupling strength varies significantly across space and time within a single dataset. Business streets exhibit strong spatial dependencies as congestion cascades upstream within minutes~\cite{xiong2018predicting}, while highway interchanges follow predictable temporal patterns~\cite{kerner2002empirical}. Similarly, rush hours demonstrate temporal regularity as traffic converges on major routes, whereas accidents trigger spatial propagation regardless of historical patterns. \textbf{(3) Suboptimal Spatial Modeling.}
Effective coupling requires accurately modeling each correlation. To capture the spatial topology, most existing approaches use either fixed pre-defined graphs that cannot capture evolving relationships~\cite{lan2022dstagnn}, or separately learned graphs that can be noisy when misestimated~\cite{kang2019robust}. Others employ attention over all location pairs, but but recomputing pairwise scores for every sample incurs substantial overhead, and spurious
long-range links harm robustness~\cite{wang2022kvt}. These limitations demand more reliable spatial modeling.

We propose \textbf{AdaST}, a simple but effective framework that adaptively modulates spatial and temporal modeling according to data-dependent coupling structures. Rather than uniformly enforcing joint spatial-temporal modeling, AdaST follows a \emph{decompose--recompose} principle: disentangling different correlation patterns can help prevent confounding interactions in adaptive coupling, which might otherwise obscure each component's contribution and induce negative transfer~\cite{xia2023deciphering, deng2024disentangling}.
Specifically, AdaST employs a decomposer that factorizes the input into three components: (i) a temporal-specific component capturing intra-series dynamics, (ii) a spatial-specific component capturing cross-location dependencies, and (iii) a spatiotemporal-coupling component capturing joint interactions. Each component is then processed by a role-aligned module -- temporal-only, spatial-only, or joint spatiotemporal modeling, respectively. Finally, an adaptive recomposer learns data-dependent mixture weights to recombine these components for prediction, while a correlation-informed modulation further reduces spurious correlations within each component.
This design allows the model to automatically allocate capacity: favoring temporal cues when history dominates, spatial cues when propagation dominates, and joint cues when both are salient.

To accommodate heterogeneous coupling dynamics, AdaST employs heterogeneity-aware experts that capture spatial and temporal variations through learned embeddings, guiding decomposition based on location-specific and period-specific characteristics.
For spatial modeling, a lightweight spatial mixer efficiently mixes features along the spatial dimension. This design offers a balanced trade-off between expressiveness and robustness, enabling global interaction while avoiding excessive computational overhead and unstable long-range dependencies.
Across diverse benchmarks, AdaST consistently outperforms all baselines. Further analysis demonstrates that AdaST provides excellent interpretability, making it a transparent and adaptable solution for various forecasting scenarios.

\vspace{-10pt}
\section{Preliminaries}
\vspace{-8pt}

\subsection{Spatial-Temporal Forecasting}
Spatial-temporal forecasting aims to predict future values across multiple locations based on historical observations that exhibit both spatial and temporal dependencies. Formally, let $\mathcal{V}$ denote a set of $N$ nodes representing spatial locations, and let $\mathbf{x}_i \in \mathbb{R}^{T \times D}$ be the time series associated with each node $v_i \in \mathcal{V}$, where $T$ is the number of historical time steps and $D$ is the feature dimensionality. The complete spatial-temporal data is represented as a tensor $\mathbf{X} = [\mathbf{x}_1, \ldots, \mathbf{x}_N]^\top \in \mathbb{R}^{N \times T \times D}$.
The task is to learn a mapping that predicts future observations $\hat{\mathbf{Y}} \in \mathbb{R}^{N \times T' \times D}$ over the next $T'$ time steps:
\begin{equation}
    \hat{\mathbf{Y}} = f(\mathbf{X}; \mathbf{A}^*),
\end{equation}
where $\mathbf{A}^* \in \{\mathbf{A}, \emptyset\}$ denotes an optional adjacency matrix $\mathbf{A} \in \mathbb{R}^{N \times N}$ encoding pre-existing spatial connectivity when prior structural knowledge is available.

\vspace{-5pt}
\subsection{Coupling Structure}
\vspace{-5pt}
We characterize how spatial and temporal dependencies interact by distinguishing coupling regimes.

\noindent\textbf{Weakly Coupled Temporal-Dominated.}
Future values at location $i$ are governed primarily by its own history, with negligible cross-location influence:
\begin{equation}
    y_i^t \approx g_{\text{temp}}\!\big(\mathbf{x}_i^{1:t-1}\big),
\end{equation}
where $g_{\text{temp}}$ models intra-series dynamics.

\noindent\textbf{Weakly Coupled Spatial-Dominated.}
Future values are driven by short-horizon propagation from other nodes, while long-term temporal trends contribute minimally:
\begin{equation}
    y_i^t \approx g_{\text{spat}}\!\big(\{\mathbf{x}_j^{t-1}\}_{j \in \mathcal{V}}\big),
\end{equation}
where $g_{\text{spat}}$ aggregates other nodes' most recent states.

\noindent\textbf{Strongly Coupled.}
Spatial and temporal effects are tightly intertwined and must be modeled jointly:
\begin{equation}
    y_i^t \approx g_{\text{st}}\!\big(\mathbf{x}_i^{1:t-1},\, \{\mathbf{x}_j^{1:t-1}\}_{j \in \mathcal{V}}\big),
\end{equation}
where $g_{\text{st}}$ integrates histories across space and time to capture their mutual dependence.

Different datasets exhibit different coupling regimes, but many existing methods apply a single, strongly-coupled spatial-temporal architecture across the board. This \emph{mismatch} between the model's inductive bias and the data's coupling pattern encourages the models to exploit incidental cross-branch associations that are spurious and not generalizable, leading to suboptimal performance.

\begin{wraptable}{r}{0.6\linewidth}
    \centering
    \small
    \vspace{-12pt}
    \caption{Average temporal and spatial correlation coefficient in three synthetic datasets. Bold values indicate dominant correlations matching the intended coupling regime.}
    \label{tab:corr}
    \begin{tabular}{lccc}
    \toprule
    \textbf{Correlation} & \textbf{TD} & \textbf{SD} & \textbf{SC}  \\
    \midrule
    Temporal   & $\textbf{0.63}\pm0.07$ & $0.25\pm0.01$ & $0.54\pm0.05$  \\
    Spatial  & $-0.05\pm0.06$ & $\textbf{0.33}\pm0.12$ & $0.21\pm0.02$ \\
    \bottomrule
    \end{tabular}
    \vspace{-8pt}
\end{wraptable}
\vspace{-5pt}
\subsection{Preliminary Experiments}
\vspace{-5pt}
\label{sec:prelim}
To empirically validate the importance of architecture-structure alignment, we conduct controlled experiments on synthetic datasets with known coupling structures.

\noindent\textbf{Synthetic Data Generation.} We construct three datasets corresponding to the three coupling structures:

\begin{itemize}
    \item \textbf{Temporal-Dominated (TD):} For each time series $i$, future values are generated solely from its own history. Each follows an order-12 Auto-Regressive process (AR(12)) with coefficients $\{\phi_{i,k}\}_{k=1}^{12}$ and an exogenous multi-frequency signal $s_i(t)$:
    \begin{equation*}
        y_i^t=\sum_{k=1}^{12}\phi_{i,k}\,\mathbf{x}_i^{t-k}+0.4\,s_i(t)+\epsilon_i^t,\quad
        \epsilon_i^t\sim\mathcal{N}(0,0.15^2).
    \end{equation*}
    \item \textbf{Spatial-Dominated (SD):} Next-step values depend only on neighbors' current states. We build a directed sparse random graph $\mathbf{A}$ (no self-loops) and ensure each node has at least one in- and out-edge. Let $P=D_{\text{out}}^{-1}\mathbf{A}$ where $D_{\text{out}}=\mathrm{diag}(\mathbf{A}\mathbf 1)$:
    \begin{equation*}
        y_i^t = \sum_{j=1}^{N} P_{ji}\,\mathbf{x}_j^{t-1}+\epsilon_i^t,\quad
        \epsilon_i^t\sim\mathcal{N}(0, 0.3^2).
    \end{equation*}
    \item \textbf{Strongly-Coupled (SC):} Future values depend on history and spatial neighbors, combining per-node AR(6) with graph diffusion and a weak exogenous term with $\epsilon_i^t \sim \mathcal{N}(0,0.15^2)$:
    \begin{align*}
        y_i^t
        = \sum_{k=1}^{6}\phi_{i,k}\,\mathbf{x}_i^{t-k}
        + \sum^N_{j=1} P_{ji}\,\mathbf{x}_j^{t-1}
        + 0.2\,s_i(t)
        + \epsilon_i^t.
    \end{align*}
\end{itemize}

We computed the correlation coefficients for the three synthetic datasets and observed that each synthetic dataset exhibits its intended coupling structure in Table~\ref{tab:corr}. Details are provided in Appendix~\ref{app:toy_data}.

\begin{wrapfigure}{r}{0.44\linewidth}
    \centering
    \vspace{-5pt}
    \includegraphics[width=\linewidth]{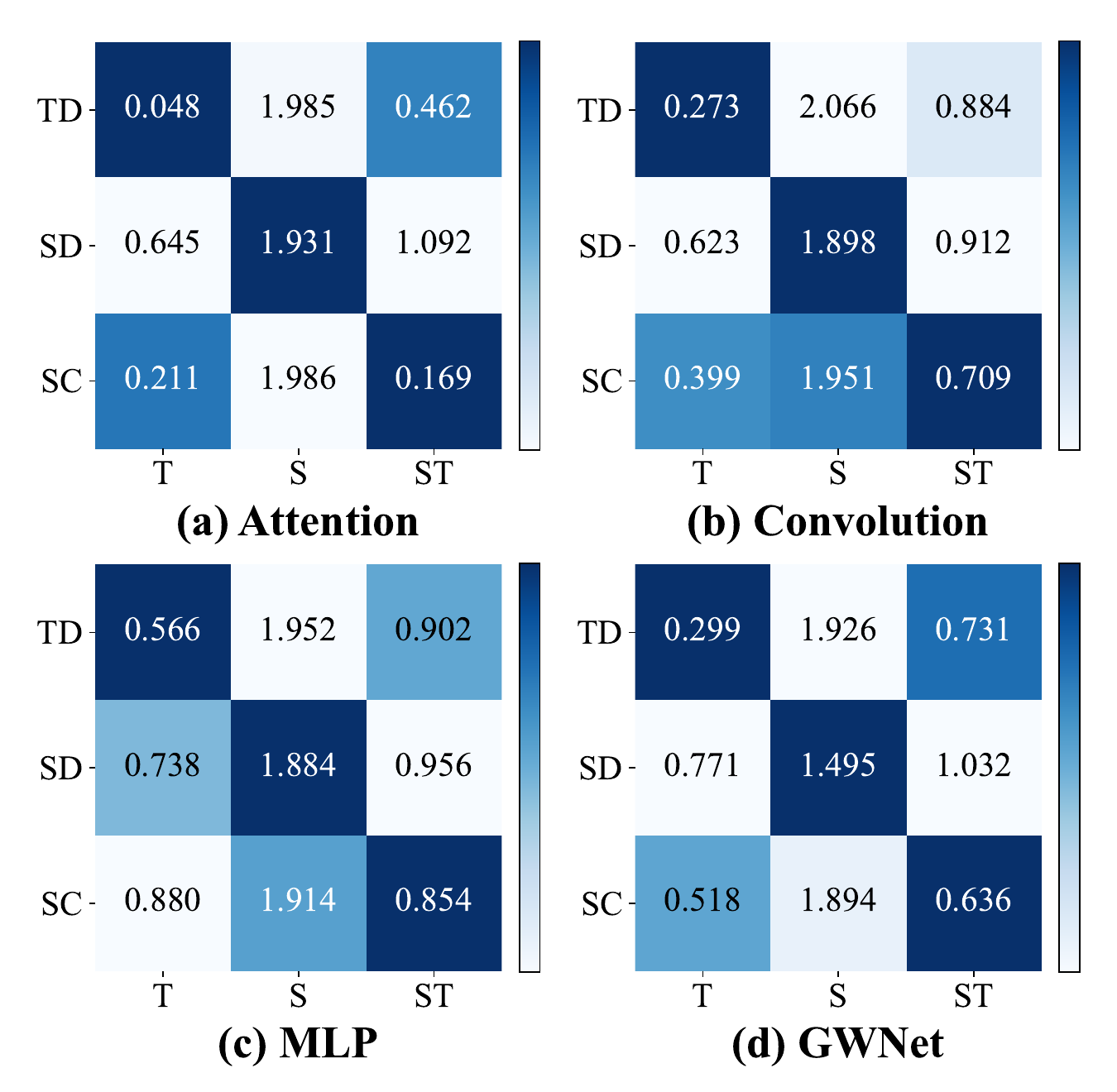}
    \caption{Performance of different architectures on synthetic datasets with known coupling structures. Darker colors indicate lower normalized MAE.}
    \label{fig:prelim}
    \vspace{-10pt}
\end{wrapfigure}

\noindent\textbf{Architecture Categorization.} We evaluate three architecture types with four backbone implementations:
\begin{itemize}
    \item \textbf{Temporal-Only (T)} models intra-series dependencies from historical values at each location independently, with no cross-location information exchange.
    \begin{equation}
        \hat{y}_i^t = f_{\text{temp}}\left(\mathbf{x}_i^{1:t-1}\right),
    \end{equation}

    \item \textbf{Spatial-Only (S)} aggregates information from other nodes at each time step through spatial convolution or attention, with simply averaging along the temporal dimension to suppress temporal pattern extraction.
    \begin{equation}
        \hat{y}_i^t = \frac{1}{T}\sum_{t=1}^{T} f_{\text{spat}}\left(\left\{\mathbf{x}_j^{t}\right\}_{j \in \mathcal{V}}, \mathbf{x}_i^{t}\right),
    \end{equation}

    \item \textbf{Spatial-Temporal (ST)} jointly models spatial and temporal correlations through coupled architectures.
    \begin{equation}
        \hat{y}_i^t = f_{\text{temp}}\left(f_{\text{spat}}\left(\left\{\mathbf{x}_j^{t}\right\}_{j \in \mathcal{V}}, \mathbf{x}_i^{t}\right)\right),
    \end{equation}
\end{itemize}
For each architecture type, we implement four backbone variants: attention-based, convolutional, MLP-based, and GWNet~\cite{wu2019graph}, a representative spatial-temporal forecasting framework.

\noindent\textbf{Experimental Results.} Figure~\ref{fig:prelim} presents the normalized Mean Absolute Error (MAE) across all architecture-data combinations. Each heatmap corresponds to one backbone type, where the x-axis represents data types (TD, SD, SC) and the y-axis represents architecture types (T, S, ST). Colors are row-normalized for visualization clarity, with darker colors indicating better performance.
From the figure, we could observe that diagonal entries consistently exhibit the darkest colors across all backbones and off-diagonal entries show significant performance degradation, demonstrating that architecture-structure alignment yields optimal performance (e.g., temporal-only architectures excel on temporal-dominated data). In addition, more expressive backbones (e.g., attention vs. convolution) exhibit larger performance gaps for mismatched architectures, suggesting that complex models are more susceptible to overfitting spurious correlations when the architecture does not align with the underlying coupling structure.
These findings motivate the need for adaptive frameworks that can dynamically modulate spatial and temporal modeling based on the inherent coupling structure of the data, rather than enforcing fixed architectural priors.

\vspace{-5pt}
\section{Methodology}
\begin{figure}[t]
    \centering
    \includegraphics[width=\linewidth]{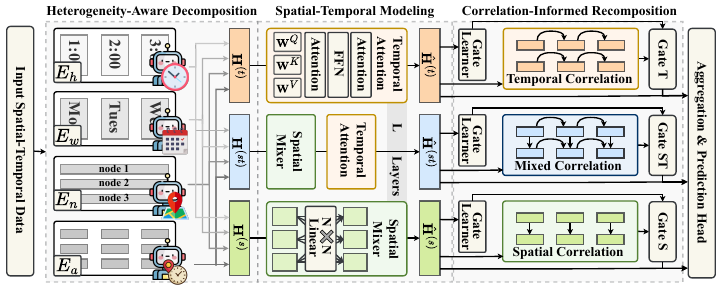}
    \vspace{-10pt}
    \caption{The overall framework of AdaST. Different colors denote different components.}
    \vspace{-10pt}
    \label{fig:overview}
\end{figure}

\vspace{-5pt}
In this section, we elaborate our methodology AdaST, which is a simple and effective framework for adaptively coupling spatial and temporal signals in forecasting. AdaST is decomposed into three modules: (1) Heterogeneity-Aware Decoupling, which utilizes several heterogeneity-aware experts to provide different views for decoupling. (2) Spatial-Temporal Modeling, which models temporal correlation with attention and spatial correlation with spatial mixer framework. (3) Correlation-Regularized Recoupling, which adaptively recouples different components with gated neural network with an correlation-aware regularization to remove spurious correlation. The overview is in Figure~\ref{fig:overview}.
\vspace{-5pt}
\subsection{Heterogeneity-Aware Decomposition}
\vspace{-5pt}
To disentangle confounding cross-pattern interactions, we decompose the input spatial-temporal data into three components: a {temporal-specific component} $\mathbf{X}^{(t)}$, a {spatial-specific component} $\mathbf{X}^{(s)}$, and a {spatiotemporal-coupling component} $\mathbf{X}^{(st)}$. Each component will then be processed by its corresponding role-aligned modeling module. However, coupling structures exhibit considerable heterogeneity across spatial locations and temporal periods. A fixed, uniform decomposition mechanism would fail to capture such variations, as it imposes identical decomposition patterns regardless of local characteristics. To address this limitation, we propose a heterogeneity-aware expert architecture wherein multiple specialized experts provide distinct decomposition perspectives that adapt to the underlying heterogeneity in coupling structures.

We introduce four expert types that provide heterogeneous decomposition perspectives: (1) \textit{Spatial Expert} for spatial heterogeneity, (2) \textit{Time-of-Day} and (3) \textit{Day-of-Week Experts} for temporal heterogeneity at different granularities, and (4) \textit{Spatiotemporal Expert} for joint variations. Each maintains learnable contextual embeddings $\mathbf{E}_n \in \mathbb{R}^{N \times D_n}$ for spatial, $\mathbf{E}_h \in \mathbb{R}^{T_{\text{day}} \times D_h}$ and $\mathbf{E}_w \in \mathbb{R}^{T_{\text{week}} \times D_w}$ for temporal, and $\mathbf{E}_a \in \mathbb{R}^{T \times N \times D_a}$ for spatial-temporal contexts, encoding time and location-specific attributes. Given input $\mathbf{X}$, we first project it to an embedding space $\mathbf{H}_0 = \text{Linear}(\mathbf{X}) \in \mathbb{R}^{T \times N \times D_0}$. Each expert $k$ then constructs contextualized representations by concatenating $\mathbf{H}_0$ with its embeddings. For the spatial and spatiotemporal experts, we have:
\begin{align}
    \mathbf{H}_n &= [\mathbf{H}_0 \| \tilde{\mathbf{E}}_n], \quad \mathbf{H}_a = [\mathbf{H}_0 \| \mathbf{E}_a],
\end{align}
where $\tilde{\mathbf{E}}_n$ denotes $\mathbf{E}_n$ expanded along temporal dimensions. For temporal experts, we first index embeddings by timestamp. Let $\mathbf{t}, \mathbf{d} \in \mathbb{R}^{T \times N}$ denote time-of-day and day-of-week indices:
\begin{align}
    \mathbf{H}_h &= [\mathbf{H}_0 \| \tilde{\mathbf{E}}_h[\mathbf{t}]], \quad \mathbf{H}_w = [\mathbf{H}_0 \| \tilde{\mathbf{E}}_w[\mathbf{d}]].
\end{align}
where $\tilde{\mathbf{E}}_h$ and $\tilde{\mathbf{E}}_w$ are expanded along spatial dimensions. Each expert then applies a projection head $f_k: \mathbb{R}^{D_0 + D_k} \rightarrow \mathbb{R}^{3/4D_H}$ to extract three components:
\begin{equation}
    [\mathbf{Z}_k^{(st)}; \mathbf{Z}_k^{(t)}; \mathbf{Z}_k^{(s)}] = f_k(\mathbf{H}_k),
\end{equation}
where $k \in \{n, h, w, a\}$ and each $\mathbf{Z}_k^{(\cdot)} \in \mathbb{R}^{T \times N \times 1/4D_H}$. Finally, we integrate perspectives from all experts via concatenation for each component:
\begin{equation}
    \mathbf{H}^{(\cdot)} = [\mathbf{Z}_n^{(\cdot)} \| \mathbf{Z}_h^{(\cdot)} \| \mathbf{Z}_w^{(\cdot)} \| \mathbf{Z}_a^{(\cdot)}] \in \mathbb{R}^{T \times N \times D_H},
\end{equation}
where $(\cdot) \in \{st, t, s\}$. The resulting $\mathbf{H}^{(st)}, \mathbf{H}^{(t)}, \mathbf{H}^{(s)}$ encode heterogeneity-aware coupling patterns and serve as inputs to subsequent role-aligned modeling modules.

\subsection{Spatial-Temporal Modeling}

After decomposition, each component is processed by role-aligned modules specialized for its correlation pattern. The temporal-specific component $\mathbf{X}^{(t)}$ is processed by temporal attention layers, the spatial-specific component $\mathbf{X}^{(s)}$ by spatial mixer layers, and the spatiotemporal-coupling component $\mathbf{X}^{(st)}$ by interleaved temporal and spatial operations.

For temporal modeling, we employ multi-head self-attention to capture long-range dependencies: 
\begin{align}
    &\mathbf{H}' = \text{MHA}(\mathbf{H}\mathbf{W}^Q, \mathbf{H}\mathbf{W}^K, \mathbf{H}\mathbf{W}^V), \\
    &\text{TempAttn}(\mathbf{H}) = \text{FFN}(\text{LN}(\mathbf{H} + \text{Dropout}(\mathbf{H}'))),
\end{align}
where LN denotes layer normalization and FFN denotes a feed-forward network. For spatial modeling, we adopt a spatial mixer with learnable assignment matrix $\mathbf{A} \in \mathbb{R}^{N \times N}$:
\begin{equation}
    \text{SpatMix}(\mathbf{H}) = \text{FFN}(\mathbf{H} + (\text{softmax}(\mathbf{A}) \mathbf{H}^{\top})^{\top}),
\end{equation}
where $\mathbf{H}^{\top} \in \mathbb{R}^{N \times (T \times D)}$ denotes the transpose along the spatial dimension. This design captures all pairwise spatial interactions without a predefined graph and substantial computation overhead.

We stack $L$ temporal attention layers for $\mathbf{H}^{(t)}$ and $L$ spatial mixer layers for $\mathbf{H}^{(s)}$. For $\mathbf{H}^{(st)}$, we interleave temporal attention and spatial mixing in each of the $L$ layers. The final outputs are $\hat{\mathbf{H}}^{(t)}, \hat{\mathbf{H}}^{(s)}, \hat{\mathbf{H}}^{(st)} \in \mathbb{R}^{T \times N \times D_H}$.

\subsection{Correlation-Informed Recomposition}

After role-aligned modeling, we obtain three processed components $\hat{\mathbf{H}}^{(t)}, \hat{\mathbf{H}}^{(s)}, \hat{\mathbf{H}}^{(st)}$ that capture distinct correlation patterns. To generate final predictions, we adaptively recombine these components through a gated aggregation mechanism that learns data-dependent mixture weights based on component representations and their correlation strengths.

We employ a lightweight gated neural network that computes adaptive weights for each component. For each component $(\cdot) \in \{t, s, st\}$, the gate is computed as:
\begin{equation}
    g^{(\cdot)} = \sigma(W_{(\cdot)} \cdot \hat{\mathbf{H}}^{(\cdot)}) \in \mathbb{R}^{T \times N},
\end{equation}
where $\sigma(\cdot)$ is the sigmoid function, $W_{(\cdot)} \in \mathbb{R}^{D_H \times 1}$ are learnable parameters. This allows the model to automatically adjust component contributions based on input characteristics and learned patterns.

To further enhance the gating mechanism, we incorporate correlation-informed weights that reflect the strength of inherent correlations within each component. The rationale is that when a component exhibits weak correlations, spurious noise dominates over meaningful signals, and excessive reliance on such components degrades predictions. We use cosine similarity as our correlation measures. For the temporal component, we measure along the temporal dimension:
\begin{equation}
    \mathcal{C}_t(\mathbf{H}^{(t)}) = \frac{1}{T-1} \sum_{i=1}^{T-1} \text{CosineSim}(\hat{\mathbf{H}}^{(t)}_{:,i,:,:}, \hat{\mathbf{H}}^{(t)}_{:,i+1,:,:}),
\end{equation}
and for the spatial component, along the spatial dimension:
\begin{equation}
    \mathcal{C}_s(\mathbf{H}^{(s)}) = \frac{1}{N-1} \sum_{j=1}^{N-1} \text{CosineSim}(\hat{\mathbf{H}}^{(s)}_{:,:,j,:}, \hat{\mathbf{H}}^{(s)}_{:,:,j+1,:}).
\end{equation}
For the spatiotemporal component, we average both:
\begin{equation}
    \mathcal{C}_{st}(\mathbf{H}^{(st)}) = \frac{1}{2}\big(\mathcal{C}_s(\mathbf{H}^{(st)}) + \mathcal{C}_t(\mathbf{H}^{(st)})\big).
\end{equation}
We then modulate the gate scores by correlation measures:
\begin{equation}
    \tilde{g}^{(\cdot)} = g^{(\cdot)} *   [\mathcal{C}_{(\cdot)}(\mathbf{H}^{(\cdot)})]^\alpha, \quad \text{for } (\cdot) \in \{t, s, st\},
\end{equation}
where $\alpha$ controls the influence of correlation strength. The final weights are obtained via softmax:
\begin{equation}
    w^{(t)}, w^{(s)}, w^{(st)} = \text{softmax}(\tilde{g}^{(t)}, \tilde{g}^{(s)}, \tilde{g}^{(st)}),
\end{equation}
and the recomposed representation is:
\begin{equation}
    \hat{\mathbf{H}} = w^{(t)} \odot \hat{\mathbf{H}}^{(t)} + w^{(s)} \odot \hat{\mathbf{H}}^{(s)} + w^{(st)} \odot \hat{\mathbf{H}}^{(st)},
\end{equation}
where $\odot$ denotes element-wise multiplication with broadcasting. Finally, we apply a temporal projection followed by an output layer to generate predictions $\hat{\mathbf{Y}}$.

\section{Experiments}
We empirically evaluate AdaST and organize this section around five research questions:
\textbf{RQ1.} How does AdaST compare with state-of-the-art baselines across diverse benchmarks?
\textbf{RQ2.} What is the contribution of each module to overall component?
\textbf{RQ3.} How do different datasets manifest distinct coupling structures?
\textbf{RQ4.} How does the coupling structure evolve over time and vary across locations?
\textbf{RQ5.} How well are coupling components disentangled?

\begin{table}[t]
\centering
\caption{Main results on PurpleAir and PEMS benchmarks. The best and second-best scores are highlighted in \textbf{bold} and \underline{underlined}. AdaST achieves the best performance across all datasets.}
\label{tab:main}
\resizebox{\textwidth}{!}{
\begin{tabular}{lcccccccccccc}
\toprule
\multirow{2}{*}{\textbf{Dataset}} & \multicolumn{3}{c}{\textbf{PurpleAir}} & \multicolumn{3}{c}{\textbf{PEMS04}} & \multicolumn{3}{c}{\textbf{PEMS07}} & \multicolumn{3}{c}{\textbf{PEMS08}} \\
\cmidrule(lr){2-4}\cmidrule(lr){5-7}\cmidrule(lr){8-10}\cmidrule(lr){11-13}
 & MAE & RMSE & MAPE & MAE & RMSE & MAPE & MAE & RMSE & MAPE & MAE & RMSE & MAPE \\
\midrule
HI & $3.430$ & $5.983$ & $72.52\%$ & $42.35$ & $61.66$ & $29.92\%$ & $49.03$ & $71.18$ & $22.75\%$ & $36.66$ & $50.45$ & $21.63\%$ \\
DeepAR & $0.994$ & $1.817$ & $32.48\%$ & $20.64$ & $32.35$ & $14.28\%$ & $22.00$ & $35.44$ & $9.31\%$ & $16.80$ & $26.38$ & $10.66\%$ \\
NBeats & $\underline{0.511}$ & $1.100$ & $23.30\%$ & $25.30$ & $39.65$ & $17.66\%$ & $26.14$ & $42.72$ & $11.37\%$ & $18.90$ & $31.39$ & $12.11\%$ \\
\midrule
GWNet & $0.514$ & $\underline{1.012}$ & $23.51\%$ & $18.80$ & $30.14$ & $13.19\%$ & $20.47$ & $33.47$ & $8.61\%$ & $14.67$ & $23.55$ & $9.46\%$ \\
DCRNN & $0.656$ & $1.268$ & $26.99\%$ & $19.63$ & $31.26$ & $13.59\%$ & $21.16$ & $34.14$ & $9.02\%$ & $15.22$ & $24.17$ & $10.21\%$ \\
AGCRN & $0.774$ & $1.595$ & $27.90\%$ & $19.38$ & $31.25$ & $13.40\%$ & $20.57$ & $34.40$ & $8.74\%$ & $15.32$ & $24.41$ & $10.03\%$ \\
STGCN & $0.619$ & $1.152$ & $27.29\%$ & $19.57$ & $31.38$ & $13.44\%$ & $21.74$ & $35.27$ & $9.24\%$ & $16.08$ & $25.39$ & $10.60\%$ \\
GTS & $0.617$ & $1.157$ & $29.90\%$ & $20.96$ & $32.95$ & $14.66\%$ & $22.15$ & $35.10$ & $9.38\%$ & $16.49$ & $26.08$ & $10.54\%$ \\
MTGNN & $0.540$ & $1.089$ & $\underline{22.65\%}$ & $19.17$ & $31.70$ & $13.37\%$ & $20.89$ & $34.06$ & $9.00\%$ & $15.18$ & $24.24$ & $10.20\%$ \\
StemGNN & $0.512$ & $1.038$ & $22.89\%$ & $22.98$ & $36.00$ & $16.56\%$ & $22.50$ & $36.41$ & $9.57\%$ & $16.90$ & $26.30$ & $11.89\%$ \\
STNorm & $0.574$ & $1.102$ & $23.69\%$ & $18.96$ & $30.98$ & $12.69\%$ & $20.50$ & $34.66$ & $8.75\%$ & $15.41$ & $24.77$ & $9.76\%$ \\
D$^2$STGNN & $0.618$ & $1.495$ & $22.71\%$ & $\underline{18.32}$ & $\underline{29.89}$ & $12.51\%$ & $19.49$ & $\underline{32.59}$ & $\underline{8.09}\%$ & $14.10$ & $23.36$ & $9.33\%$ \\
STID & $0.563$ & $1.137$ & $26.69\%$ & $18.42$ & $29.91$ & $12.39\%$ & $19.61$ & $32.69$ & $8.31\%$ & $14.21$ & $23.35$ & $9.32\%$ \\
STDN & $0.672$ & $1.223$ & $29.23\%$ & $18.40$ & $30.41$ & $\underline{12.21\%}$ & $20.08$ & $33.73$ & $9.29\%$ & $14.21$ & $\underline{23.28}$ & $9.27\%$ \\
HimNet & $0.574$ & $1.139$ & $24.43\%$ & $\underline{18.32}$ & $30.07$ & $12.49\%$ & $20.03$ & $33.17$ & $8.79\%$ & $\underline{13.52}$ & $\underline{23.28}$ & $8.96\%$ \\
STAEformer & $0.563$ & $1.060$ & $22.98\%$ & $18.35$ & $30.73$ & $12.25\%$ & $\underline{19.46}$ & $33.56$ & $8.41\%$ & $13.56$ & $23.72$ & $\underline{8.92}\%$ \\
\midrule
\textbf{AdaST} & $\textbf{0.489}$ & $\textbf{0.994}$ & $\textbf{22.43\%}$ & $\textbf{18.28}$ & $\textbf{29.87}$ & $\textbf{12.17\%}$ & $\textbf{19.16}$ & $\textbf{32.54}$ & $\textbf{8.02}\%$ & $\textbf{13.45}$ & $\textbf{23.17}$ & $\textbf{8.85}\%$ \\
\bottomrule
\end{tabular}
}
\vspace{-10pt}
\end{table}

\subsection{Experimental Settings}
\vspace{-10pt}
\noindent\textbf{Datasets.}
We evaluate AdaST on four spatial-temporal forecasting benchmarks with diverse coupling characteristics. PEMS04, PEMS07, and PEMS08 are traffic flow datasets from the California Transportation Performance Management System, sampled at 5-minute intervals ($T_\text{day}=288$ time steps per day)~\cite{song2020spatial}. PurpleAir contains PM2.5 measurements from air quality sensors in the Boston area. We preprocess the raw 2-minute data by resampling to 6-minute intervals to mitigate noise and missing values ($T_\text{day}=240$ time steps per day). All datasets use weekly periodicity with $T_\text{week}=7$. Following standard practice, we apply Z-score normalization and split each dataset into 60\% training, 20\% validation, and 20\% testing.

\noindent\textbf{Baselines.}
We compare AdaST against 16 representative methods. Temporal-only baselines include HI~\cite{cui2021historical}, DeepAR~\cite{salinas2020deepar}, and NBeats~\cite{oreshkin2019n}. Spatial-temporal baselines include GWNet~\cite{wu2019graph}, DCRNN~\cite{li2017diffusion}, AGCRN~\cite{bai2020adaptive}, STGCN~\cite{yu2017spatio}, GTS~\cite{shang2021discrete}, MTGNN~\cite{wu2020connecting}, StemGNN~\cite{cao2020spectral}, STNorm~\cite{deng2021st}, D$^2$STGNN~\cite{shao2022decoupled}, STID~\cite{shao2022spatial}, HimNet~\cite{dong2024heterogeneity}, STAEformer~\cite{liu2023staeformer}, and STDN~\cite{cao2025spatiotemporal}.

\noindent\textbf{Implementation Details.}
We set expert embedding dimensions to $D_n=D_h=D_w=D_a=24$ and hidden dimension to $D_H=256$. The model uses $L=3$ layers with 4-head temporal attention. Both input and prediction lengths are 12 time steps ($T=T'=12$). We train with the Adam optimizer (initial learning rate 0.001, exponential decay, batch size 16) and set the correlation weight to $\alpha=0.1$. All experiments are conducted on an NVIDIA A100 80GB GPU using official baseline implementations with recommended hyperparameters. Our code is available at \url{https://github.com/LzyFischer/AdaST}.

\noindent\textbf{Evaluation Metrics.}
We use three standard metrics: Mean Absolute Error (MAE), Root Mean Square Error (RMSE), and Mean Absolute Percentage Error (MAPE). Following convention, we report average performance across all 12 prediction horizons.

\vspace{-5pt}
\subsection{Main Results}
\vspace{-5pt}
To answer \textbf{RQ1}, we present the comprehensive comparison results in Table~\ref{tab:main}. We make the following key observations:
(1) AdaST achieves state-of-the-art results across all four benchmarks, consistently outperforming the strongest baselines. Compared to the best-performing competitors, AdaST improves MAE by up to $1.7\%,$ RMSE by $0.7\%$, and MAPE by $0.8\%$ on average, demonstrating the effectiveness of adaptive coupling mechanisms for spatial-temporal forecasting.
(2) Spatial-temporal methods generally outperform temporal-only baselines, confirming that explicitly modeling spatial dependencies is crucial for forecasting tasks where cross-location interactions exist. However, the performance gap varies significantly across datasets, suggesting heterogeneous coupling structures.
(3) AdaST's improvement is most pronounced on PurpleAir, where it outperforms the best baseline by $4.3\%$ in MAE. This larger gain can be attributed to PurpleAir's temporal-dominated coupling (as shown in Section~\ref{sec:structure}), where traditional spatial-temporal methods introduce spurious spatial correlations. AdaST's adaptive mechanism effectively suppresses unnecessary spatial modeling through lower spatial gate scores, avoiding spurious dependencies while retaining beneficial signals.

\begin{wraptable}{r}{0.55\linewidth}
    \centering
    \small
    \vspace{-12pt}
    \caption{Ablation study on PurpleAir and PEMS07. Removing each component results in performance drop.}
    \label{tab:ablation}
    \resizebox{\linewidth}{!}{
    \begin{tabular}{lcccccc}
    \toprule
    \multirow{2}{*}{\textbf{Dataset}} & \multicolumn{3}{c}{\textbf{PurpleAir}} & \multicolumn{3}{c}{\textbf{PEMS07}} \\
    \cmidrule(lr){2-4}\cmidrule(lr){5-7}
     & MAE & RMSE & MAPE & MAE & RMSE & MAPE \\
    \midrule
    {\textit{w/o} $\mathbf{E_n}$} & $0.519$ & $1.072$ & $23.79\%$ & $19.44$ & $32.84$ & $8.49\%$ \\
    {\textit{w/o} $\mathbf{E_h}$} & $\underline{0.495}$ & $\underline{0.999}$ & $\underline{22.50\%}$ & $20.39$ & $34.14$ & $8.68\%$ \\
    {\textit{w/o} $\mathbf{E_w}$} & $0.498$ & $1.023$ & $22.85\%$ & $\underline{19.21}$ & $\underline{32.63}$ & $\underline{8.07\%}$ \\
    {\textit{w/o} $\mathbf{E_a}$} & $0.516$ & $1.075$ & $23.47\%$ & $20.16$ & $33.23$ & $11.26\%$ \\
    \midrule
    \textit{SpaAtt} & $0.518$ & $1.072$ & $23.42\%$ & $19.35$ & $32.93$ & $8.11\%$ \\
    \midrule
    {\textit{w/o} $\mathcal{C}$} & $0.513$ & $1.067$ & $23.25\%$ & $19.85$ & $33.02$ & $10.58\%$ \\
    {\textit{w/o} $g$} & $0.526$ & $1.088$ & $24.00\%$ & $20.29$ & $33.35$ & $9.57\%$ \\
    \midrule
    {\textbf{AdaST}} & $\textbf{0.489}$ & $\textbf{0.994}$ & $\textbf{22.43\%}$ & $\textbf{19.16}$ & $\textbf{32.54}$ & $\textbf{8.02\%}$ \\
    \bottomrule
    \end{tabular}
    }
    \vspace{-8pt}
\end{wraptable}
\vspace{-5pt}
\subsection{Ablation Study}
\vspace{-5pt}
\label{sec:ablation}
To answer \textbf{RQ2}, we systematically ablate key modules to evaluate their individual contributions on PurpleAir and PEMS07. Table~\ref{tab:ablation} presents results for the following variants: removing individual heterogeneity experts (\textit{w/o} $\mathbf{E_n}$, \textit{w/o} $\mathbf{E_h}$, \textit{w/o} $\mathbf{E_w}$, \textit{w/o} $\mathbf{E_a}$), replacing spatial mixer with spatial attention (\textit{SpaAtt}), removing learned gate scores by using uniform averaging (\textit{w/o} $g$), and removing correlation measures from recomposition (\textit{w/o} $\mathcal{C}$).
(1) All four experts contribute to performance, though their relative importance varies across datasets. On PurpleAir, removing the spatial expert $\mathbf{E_n}$ causes the largest degradation, indicating that location-specific coupling patterns are most critical for air quality data where region characteristics differ substantially. Conversely, on PEMS07, removing the time-of-day expert $\mathbf{E_h}$ results in the greatest performance drop, reflecting the importance of diurnal traffic patterns.
(2) Replacing spatial mixer with spatial attention (\textit{SpaAtt}) degrades performance, particularly on temporal-dominated PurpleAir. This suggests that attention can introduce spurious long-range correlations when spatial coupling is weak.
(3) Both the learned gate scores and correlation measures are essential for effective recomposition. Removing gates eliminates dataset-adaptive weighting, while removing correlation measures allows spurious correlations to mislead recomposition. Together, these mechanisms enable AdaST to dynamically balance components based on their reliability and relevance.

\begin{wrapfigure}{r}{0.5\linewidth}
    \centering
    \vspace{-28pt}
    \includegraphics[width=\linewidth]{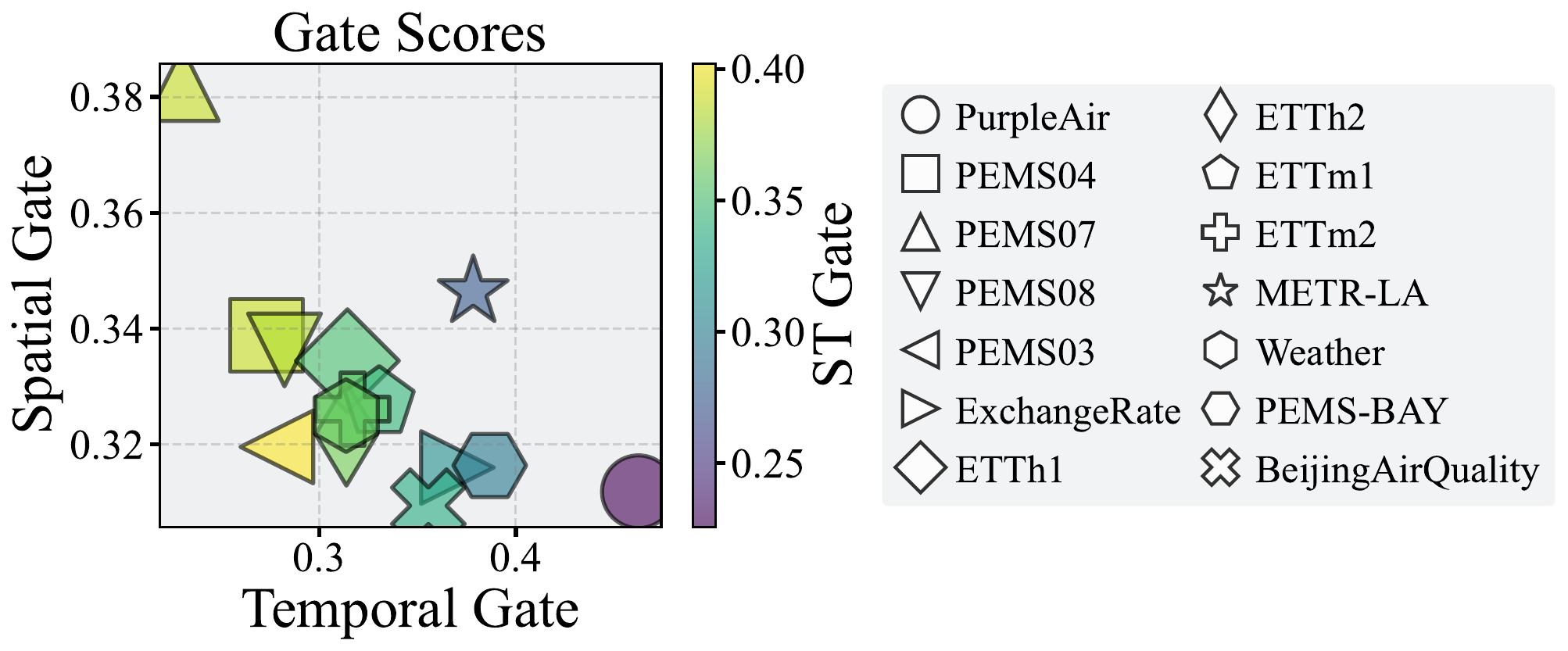}
    \caption{Average gate scores of different datasets, revealing data-specific coupling structures.}
    \label{fig:coupling_structure}
    \vspace{-8pt}
\end{wrapfigure}
\vspace{-5pt}
\subsection{Coupling Structure Analysis}
\vspace{-5pt}
\label{sec:structure}
To address \textbf{RQ3}, we investigate how coupling structures vary across common spatial-temporal datasets by visualizing the averaged gate scores $w^{(t)}$, $w^{(s)}$, and $w^{(st)}$ in Figure~\ref{fig:coupling_structure}.
We have the following observations:
(1) Different datasets exhibit distinct coupling structures, confirming that a one-size-fits-all architectural approach is suboptimal.
(2) Traditional spatial-temporal forecasting benchmarks PEMS03-PEMS08 exhibit relatively spatial-dominant patterns. This reflects the physical reality of traffic networks where congestion propagates spatially through road connections.
(3) PurpleAir and BeijingAirQuality demonstrate higher temporal gate scores, which can be attributed to the localized nature of air quality measurements, where sensor readings are primarily governed by local emission sources and meteorological conditions rather than immediate spatial diffusion from neighboring sensors. This observation aligns with our main results (Table~\ref{tab:main}), where the temporal-only baseline NBeats achieves the second-best performance.
(4) For the single-location multivariate ETTh and ETTm datasets, we repurpose the spatial module to model feature dimensions as pseudo-spatial nodes. We observe significant cross-feature coupling, evidenced by spatial gate scores comparable in magnitude to temporal ones. This confirms that inter-variate interactions provide critical predictive signals, aligning with findings on these datasets~\cite{grigsby2021long}.
(5) METR-LA and PEMS-BAY exhibit stronger temporal dominance compared to PEMS03-PEMS08. This difference may reflect variations in road network characteristics: highway systems (PEMS-BAY covers the Bay Area highway network) often exhibit more predictable temporal patterns driven by commuting schedules, while urban road networks (PEMS from various California districts) demonstrate stronger spatial propagation due to higher intersection density and more complex traffic interactions.

\vspace{-10pt}
\subsection{Case Study}
\begin{wrapfigure}{r}{0.5\linewidth}
    \centering
    \vspace{-15pt}
    \includegraphics[width=\linewidth]{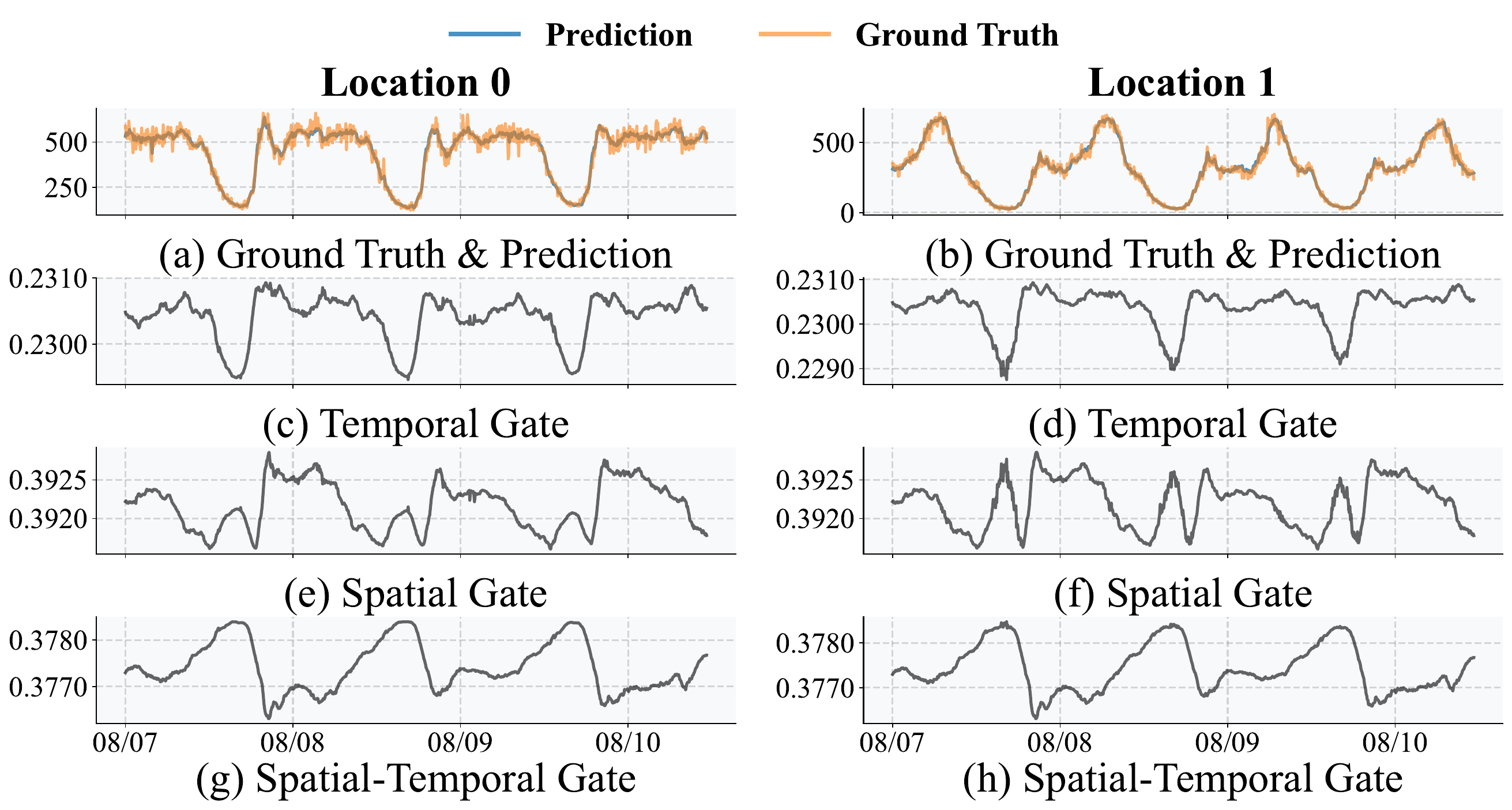}
    \caption{Gate scores across different time periods and locations, illustrating dynamic coupling.}
    \label{fig:case_study}
    \vspace{-8pt}
\end{wrapfigure}

To address \textbf{RQ4}, we present a case study on PEMS07 visualizing the learned gate scores across different time periods and locations in Figure~\ref{fig:case_study}. We observe four key patterns. (1) Gate scores are relatively stable across the whole datasets, which demonstrate the whole datasets share similar coupling structure. (2) All gate scores exhibit clear periodic patterns that align with the underlying data periodicity, demonstrating that coupling structures are temporally heterogeneous and recurrent. (3) Different spatial locations exhibit distinct coupling structures, validating the use of spatial experts that can adaptively capture location-specific dependencies. (4) Temporal gate scores peak during stable, gradual changes, while spatial gate scores spike during abrupt trend shifts. Notably, when temporal and spatial gate scores are more comparable in magnitude, the spatial-temporal gate score increases, indicating strong coupling.

\begin{wrapfigure}{r}{1.05\linewidth}
    \centering
    \vspace{-30pt}
    \includegraphics[width=\linewidth]{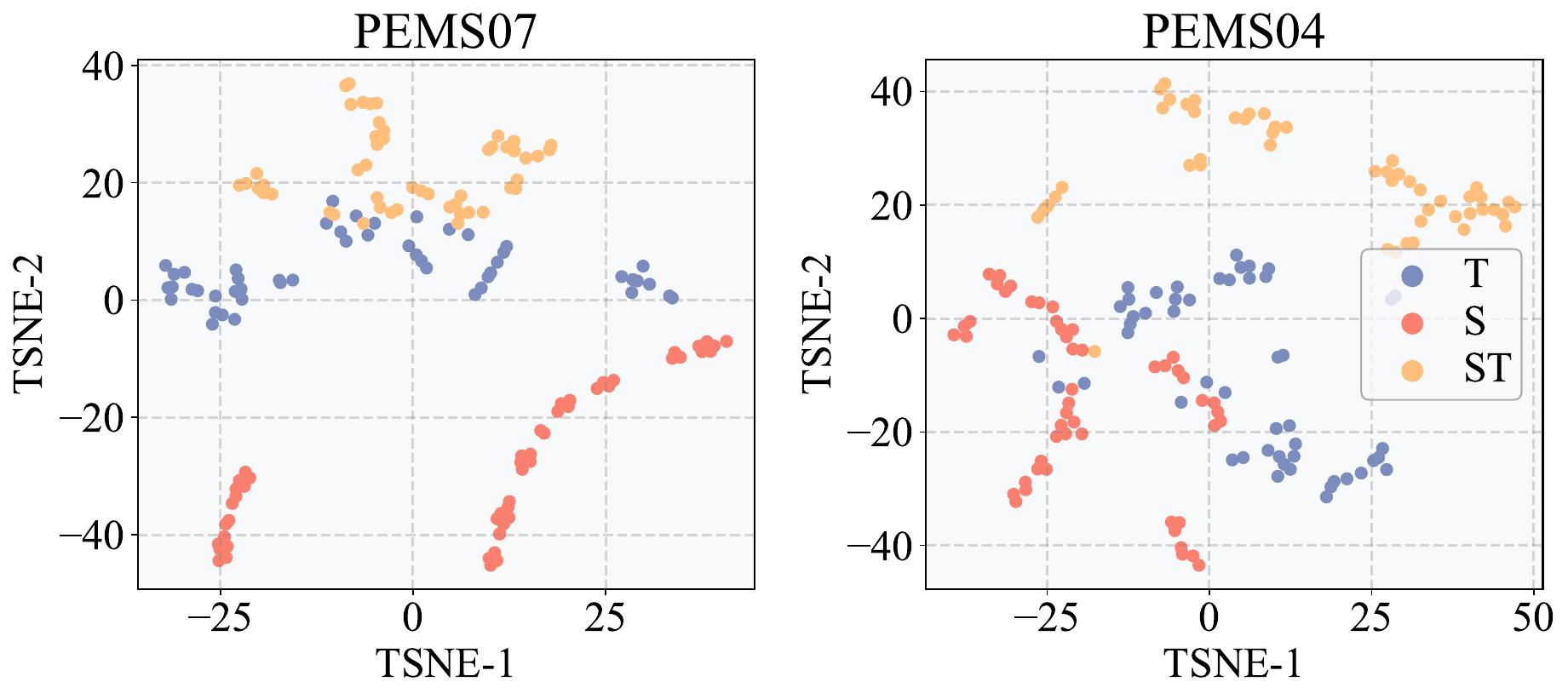}
    \caption{T-SNE visualization of learned representations for temporal (T), spatial (S), and spatial-temporal-coupling (ST) components, demonstrating clear separation and effective disentanglement.}
    \label{fig:tsne}
    \vspace{-8pt}
\end{wrapfigure}
\subsection{Representation Analysis}
To address \textbf{RQ5}, we visualize the learned representations of the three components $\hat{\mathbf{H}}^{(t)}$, $\hat{\mathbf{H}}^{(s)}$, and $\hat{\mathbf{H}}^{(st)}$ before adaptive recomposition. We apply t-SNE dimensionality reduction to project the embeddings into 2D space, as shown in Figure~\ref{fig:tsne}. The results reveal clear separation between the three component clusters, which demonstrates that our heterogeneity-aware decomposition and role-aligned modeling successfully enforce specialization, thereby providing a reliable foundation for adaptive recomposition.

\vspace{-5pt}
\section{Related Works}
\vspace{-5pt}
Spatial-temporal data encodes both temporal dynamics and spatial dependencies across multiple locations.
\textbf{Temporal-Only Approaches.}
Early forecasting methods such as ARIMA~\cite{contreras2003arima}, NBeats~\cite{oreshkin2019n}, and DeepAR~\cite{salinas2020deepar} focused on temporal modeling, treating each time series independently. While effective for intra-series dependencies, they fail to capture critical spatial correlations.
\textbf{Coupled Spatial-Temporal Models.}
To capture spatial dependencies, recent works explicitly couple temporal and spatial modeling. Graph-based methods leverage GNNs to model spatial correlations: DCRNN~\cite{li2017diffusion} combines diffusion convolution with GRUs, STGCN~\cite{yu2017spatio} stacks temporal and spatial convolutions, and GWNet~\cite{wu2019graph} introduces adaptive adjacency matrices. More recently, attention-based architectures~\cite{guo2019attention} have achieved state-of-the-art performance by capturing flexible, data-driven correlations. GMAN~\cite{zheng2020gman} employs spatial-temporal attention for long-range dependencies, while STAEformer~\cite{liu2023staeformer} introduces hierarchical attention for multi-scale modeling. These methods demonstrate substantial improvements by jointly modeling spatial and temporal correlations.
\textbf{Decoupling and Decomposition.}
Existing methods overlook that coupling structures vary significantly across datasets, leading to spurious correlations when architectural assumptions mismatch data characteristics. D$^2$STGNN~\cite{shao2022decoupled} decomposes temporal from spatial-temporal components to preserve location-specific patterns. However, it lacks a systematic decomposition of all three correlation types, and employs fixed recomposition without adaptive mechanisms.
In this work, we propose AdaST, which adaptively couple spatial and temporal signals in forecasting, enabling automatic adaptation to diverse coupling structures.

\vspace{-5pt}
\section{Conclusion}
\vspace{-5pt}
In this paper, we highlight a fundamental challenge in spatial-temporal forecasting: coupling structures vary significantly across datasets, yet existing approaches apply uniform architectural priors that fail to accommodate this variability. To address this mismatch, we propose AdaST, an adaptive framework that dynamically modulates spatial and temporal modeling through a decompose-recompose paradigm. By disentangling temporal-specific, spatial-specific, and spatiotemporally coupled patterns and adaptively recombining them in a data-dependent manner, AdaST aligns model inductive biases with intrinsic coupling structures. Extensive experiments demonstrate that AdaST consistently outperforms state-of-the-art baselines across diverse benchmarks, while providing interpretable insights into dataset-specific coupling characteristics.

\begin{ack}
The authors declare no competing interests. This work was supported in part by the National Science Foundation (NSF) under Grants 2125326, 2144209, 2223769, 2228534, 2402438, 2411248, and 2601942; the Office of Naval Research (ONR) under Grant N000142412636; the Commonwealth Cyber Initiative (CCI) under Grant HV-4Q26-073.
\end{ack}

\bibliography{neurips}
\bibliographystyle{plain}

\newpage
\appendix

\section*{Appendix Overview}
\vspace{10pt}
{\large
\begin{enumerate}[label=\Alph*., leftmargin=*, itemsep=8pt]
    \item \textbf{Synthetic Dataset Details} \dotfill \pageref{app:toy_data}
    \item \textbf{Implementation and Efficiency Analysis} \dotfill \pageref{app:impl}
    \item \textbf{Additional Main Results} \dotfill \pageref{app:additional_results}
    \item \textbf{Additional Ablation Study} \dotfill \pageref{app:additional_ablation}
    \item \textbf{Extended Interpretability Analysis} \dotfill 
    \pageref{app:interpretability}
    \item \textbf{Limitation and Broader Impact} \dotfill 
    \pageref{app:limit}
\end{enumerate}
}

\newpage

\section{Synthetic Dataset Details}
\label{app:toy_data}
In this section, we provide comprehensive details of the synthetic datasets used in Section~\ref{sec:prelim}.

\noindent\textbf{Data Generation.}
For each synthetic dataset, we generate $N=10$ spatially distributed time series, each consisting of $T=200$ time steps. The autoregressive coefficients $\{\phi_{i,k}\}_{k=1}^{12}$ are randomly sampled from a uniform distribution $\mathcal{U}(0, 0.5)$ for each series $i$, ensuring stable AR processes. The multi-frequency exogenous signals $s_i(t)$ are constructed by summing three sinusoidal components with randomly selected frequencies and amplitudes uniformly sampled from $[0, 0.5]$. For the spatial-dominated dataset, the adjacency matrix $\mathbf{A}$ is generated as a directed sparse random graph with an edge density of $0.43$, where we enforce that each node has at least one incoming and one outgoing edge to ensure connectivity.

\begin{figure}[t]
    \centering
    \includegraphics[width=\linewidth]{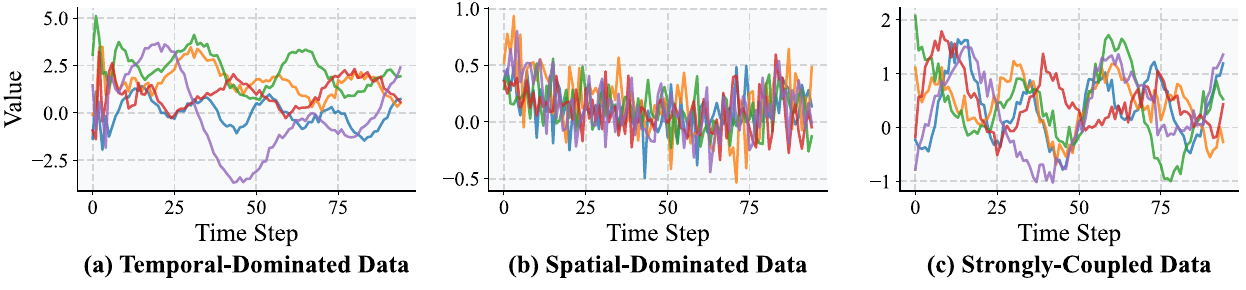}
    \caption{Visualization of three synthetic datasets with distinct coupling structures: (a) Temporal-Dominated (TD), (b) Spatial-Dominated (SD), and (c) Strongly-Coupled (SC).}
    \label{fig:toy_data}
\end{figure}

\noindent\textbf{Qualitative Analysis.}
Figure~\ref{fig:toy_data} visualizes representative time series from the three synthetic datasets. We observe distinct patterns that reflect their underlying coupling structures:
(1) The temporal-dominated (TD) dataset exhibits smooth trajectories where historical information provides reliable predictive signals with minimal noise interference from neighboring series.
(2) The spatial-dominated (SD) dataset displays more volatile, noisy patterns that prevent long-term historical information from being informative; instead, predictions rely primarily on short-term spatial information from neighbors at the most recent time step.
(3) The strongly-coupled (SC) dataset shows intermediate characteristics, combining both smooth temporal trends and spatial propagation effects.

\begin{figure}[t]
    \centering
    \includegraphics[width=\linewidth]{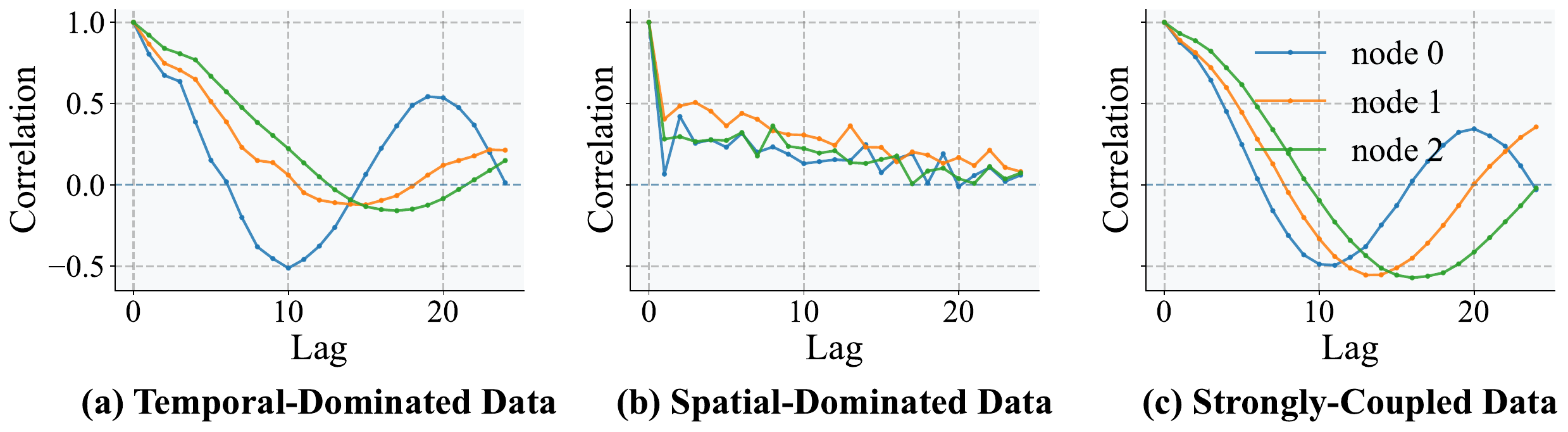}
    \caption{Temporal correlation (ACF) analysis: TD data exhibits strong autocorrelation while SD data shows weak temporal dependencies.}
    \label{fig:temcorr}
\end{figure}

\begin{figure}[t]
    \centering
    \includegraphics[width=\linewidth]{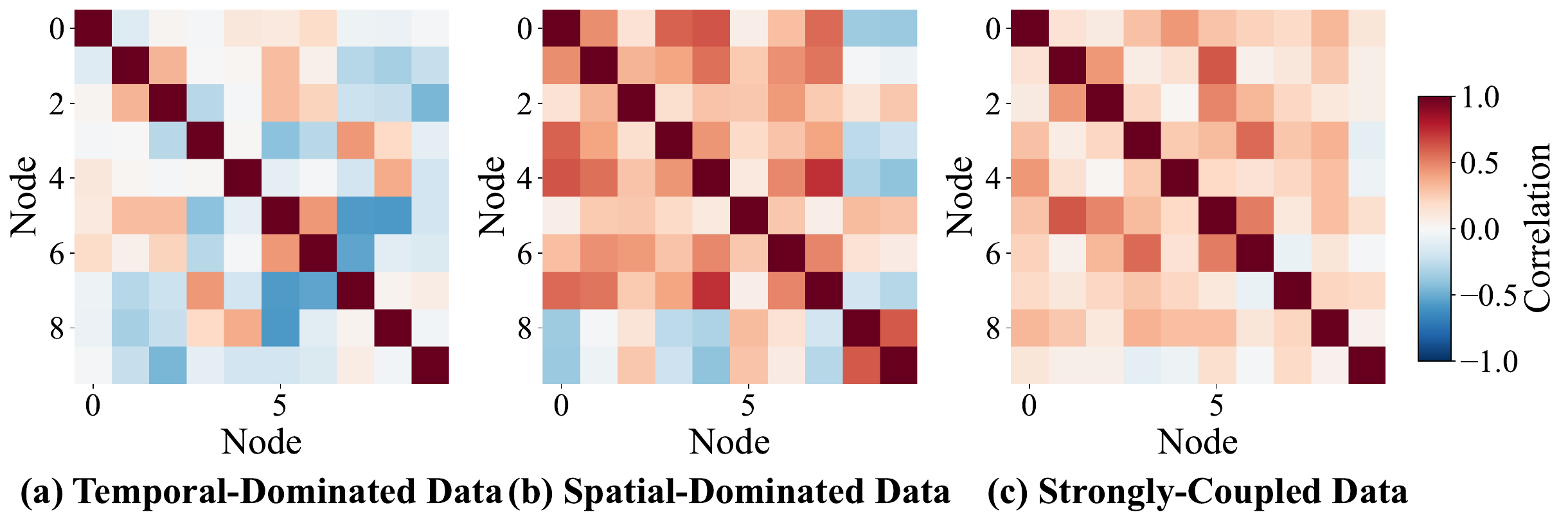}
    \caption{Spatial correlation (Pearson) analysis: SD data exhibits the strongest spatial dependencies while TD data shows negligible spatial correlation.}
    \label{fig:spacorr}
\end{figure}

\noindent\textbf{Quantitative Correlation Analysis.}
To quantitatively validate the intended coupling structures, we analyze temporal and spatial correlations in each synthetic dataset. Figure~\ref{fig:temcorr} shows the autocorrelation function (ACF) for temporal correlation, while Figure~\ref{fig:spacorr} visualizes spatial correlation measured using Pearson correlation coefficients between spatially connected node pairs.
The results confirm our design intent: temporal-dominated data exhibits significantly higher temporal correlation ($0.63\pm0.07$) but negligible spatial correlation ($-0.05\pm0.06$), while spatial-dominated data shows the opposite pattern with stronger spatial correlation ($0.33\pm0.12$) and weaker temporal correlation ($0.25\pm0.01$). The strongly-coupled dataset demonstrates moderate levels of both correlations ($0.54\pm0.05$ temporal, $0.21\pm0.02$ spatial), validating the successful construction of three distinct coupling regimes.

\newpage

\section{Implementation and Efficiency Analysis}
\label{app:impl}

\noindent\textbf{Hyperparameter Sensitivity.}
We set expert embedding dimensions to $D_n=D_h=D_w=D_a=24$ and hidden dimension to $D_H=256$. The model uses $L=3$ layers with 4-head temporal attention. The correlation modulation weight $\alpha=0.1$ is selected via grid search over $\{0.01, 0.05, 0.1, 0.5, 1.0\}$ on the validation set. Both input and prediction lengths are fixed at 12 time steps ($T=T'=12$).

\noindent\textbf{Efficiency Analysis of Spatial Mixer.}
To evaluate the computational efficiency of our spatial mixer, we compare training time per epoch against spatial attention across all four main benchmarks. As shown in Table~\ref{tab:efficiency}, the spatial mixer consistently achieves substantial speedups over spatial attention, ranging from $1.5\times$ on PEMS04 to $2.0\times$ on PurpleAir. This efficiency gain stems from the spatial mixer's input-independent mixing matrix: although both operators share the same $\mathcal{O}(N^2TD)$ aggregation cost, the spatial mixer computes $\text{softmax}(\mathbf{A})$ only once per forward pass, avoiding the per-sample query-key scoring, softmax normalization, and attention-map storage required by spatial attention. 

\begin{table}[h]
\centering
\caption{Average training time per epoch (seconds) comparing spatial mixer and spatial attention.}
\label{tab:efficiency}
\resizebox{\linewidth}{!}{
\begin{tabular}{lcccc}
\toprule
\textbf{Model} & \textbf{PEMS04} & \textbf{PEMS07} & \textbf{PEMS08} & \textbf{PurpleAir} \\
\midrule
\textit{SpaMixer} & $233$ s & $1154$ s & $141$ s & $66$ s \\
\textit{SpaAtt}   & $341$ s & $2160$ s & $220$ s & $134$ s \\
\midrule
Speedup           & $1.5\times$ & $1.9\times$ & $1.6\times$ & $2.0\times$ \\
\bottomrule
\end{tabular}
}
\end{table}

\newpage

\section{Additional Main Results}
\label{app:additional_results}

We conduct additional experiments on three diverse dataset categories to further validate AdaST's generalizability: ExchangeRate (financial), ETTh1 (energy), and METR-LA (highway traffic). We expand the baseline comparison to include two state-of-the-art time-series-only methods, PatchTST~\cite{nie2022time} and DLinear~\cite{zeng2023transformers}. For spatial-temporal baselines that require a predefined graph structure, we only compare against STID, HimNet, and STNorm, which operate without predefined graphs. Since some datasets contain large amounts of missing data and zero values that render RMSE and MAPE unstable, we report MAE as the primary metric. All experiments follow the same short-term forecasting setting as the main paper, with input and prediction lengths both fixed at 12 time steps.

Table~\ref{tab:additional} presents the results, from which we draw the following observations:
(1) AdaST achieves the best average rank across all datasets, demonstrating strong generalizability beyond the traffic and air quality domains evaluated in the main paper.
(2) Spatial-temporal baselines generally match or outperform time-series-only baselines on datasets with non-negligible spatial correlations (e.g., METR-LA, PEMS08), confirming the value of explicit spatial modeling when cross-location dependencies are present.
(3) Conversely, time-series-only baselines such as PatchTST and DLinear outperform spatial-temporal counterparts on temporally-dominated datasets such as ExchangeRate and ETTh1. This is consistent with our central argument: forcing spatial coupling onto temporally-dominated data introduces spurious cross-location dependencies that degrade predictive performance.

\begin{table}[h]
\centering
\caption{Results on additional benchmarks. Best and second-best scores are in \textbf{bold} and \underline{underlined}.}
\label{tab:additional}
\resizebox{\textwidth}{!}{
\begin{tabular}{lcccccc}
\toprule
\multirow{2}{*}{\textbf{Method}} & \textbf{ExchangeRate} & \textbf{ETTh1} & \textbf{METR-LA} & \textbf{PurpleAir} & \textbf{PEMS08} & \multirow{2}{*}{\textbf{Avg. Rank}} \\
\cmidrule(lr){2-2}\cmidrule(lr){3-3}\cmidrule(lr){4-4}\cmidrule(lr){5-5}\cmidrule(lr){6-6}
 & MAE / Rank & MAE / Rank & MAE / Rank & MAE / Rank & MAE / Rank & \\
\midrule
PatchTST       & $\textbf{0.073}$ / 1    & $0.461$ / 4             & $4.762$ / 5             & $\underline{0.505}$ / 2 & $22.07$ / 5            & 3.4 \\
DLinear        & $0.077$ / 3             & $\underline{0.444}$ / 2 & $4.820$ / 6             & $0.525$ / 3             & $22.51$ / 6            & 4.0 \\
STID           & $\underline{0.075}$ / 2 & $0.466$ / 5             & $3.146$ / 3             & $0.563$ / 4             & $14.21$ / 4             & 3.6 \\
STNorm         & $0.108$ / 5             & $0.492$ / 6             & $3.153$ / 4             & $0.574$ / 5             & $15.41$ / 3             & 4.6 \\
HimNet         & $0.124$ / 6             & $0.458$ / 3             & $\underline{3.131}$ / 2 & $0.574$ / 5             & $\underline{13.52}$ / 2 & 3.6 \\
\textbf{AdaST} & $0.083$ / 4             & $\textbf{0.407}$ / 1    & $\textbf{3.128}$ / 1    & $\textbf{0.489}$ / 1    & $\textbf{13.45}$ / 1    & 1.6 \\
\bottomrule
\end{tabular}
}
\end{table}

\newpage

\section{Additional Ablation Study}
\label{app:additional_ablation}

We present additional ablation experiments on PEMS04 and PEMS08 to complement the analysis in Section~\ref{sec:ablation}. Table~\ref{tab:ablation_app} leads to the following observations:
(1) Consistent with our findings on PurpleAir and PEMS07, all four heterogeneity-aware experts contribute meaningfully to overall performance, though their relative importance varies across datasets. On both PEMS04 and PEMS08, removing the time-of-day expert $\mathbf{E_h}$ causes the most substantial performance degradation, reflecting the critical role of diurnal traffic patterns in these datasets.
(2) Replacing the spatial mixer with spatial attention (\textit{SpaAtt}) yields only marginal differences on PEMS04 and PEMS08. This suggests that the performance gap between these two spatial modeling strategies is less pronounced on strongly spatial-coupled traffic datasets, compared to temporal-dominated data like PurpleAir where attention introduces spurious long-range correlations.
(3) Both learned gate scores ($g$) and correlation measures ($\mathcal{C}$) are essential for effective recomposition, enabling AdaST to dynamically balance component contributions based on their reliability and relevance across diverse coupling structures.

\begin{table}[h]
\centering
\caption{Ablation study on PEMS04 and PEMS08.}
\label{tab:ablation_app}
\resizebox{\linewidth}{!}{
\begin{tabular}{lcccccc}
\toprule
\multirow{2}{*}{\textbf{Variant}} & \multicolumn{3}{c}{\textbf{PEMS04}} & \multicolumn{3}{c}{\textbf{PEMS08}} \\
\cmidrule(lr){2-4}\cmidrule(lr){5-7}
 & MAE & RMSE & MAPE & MAE & RMSE & MAPE \\
\midrule
\textit{w/o} $\mathbf{E_n}$ & ${18.45}$ & ${29.93}$ & ${12.76\%}$ & ${13.68}$ & ${23.37}$ & ${9.64\%}$ \\
\textit{w/o} $\mathbf{E_h}$ & $18.58$   & $30.15$   & ${12.42\%}$ & ${14.65}$ & ${23.94}$ & ${9.51\%}$ \\
\textit{w/o} $\mathbf{E_w}$ & $\underline{18.35}$ & $\underline{29.88}$ & ${12.68\%}$ & ${13.60}$ & ${23.36}$ & ${9.25\%}$ \\
\textit{w/o} $\mathbf{E_a}$ & $18.36$   & $29.96$   & $\underline{12.23\%}$ & ${13.70}$ & ${23.70}$ & ${9.03\%}$ \\
\midrule
\textit{SpaAtt} & $\textbf{18.28}$ & $30.15$ & ${12.28\%}$ & $\underline{13.51}$ & $\underline{23.26}$ & ${9.03\%}$ \\
\midrule
\textit{w/o} $\mathcal{C}$ & $18.47$ & $30.01$ & ${12.67\%}$ & ${13.60}$ & ${23.52}$ & $\underline{8.98\%}$ \\
\textit{w/o} $g$            & $18.83$ & $30.27$ & ${13.46\%}$ & ${14.56}$ & ${23.95}$ & ${11.92\%}$ \\
\midrule
\textbf{AdaST} & $\textbf{18.28}$ & $\textbf{29.87}$ & $\textbf{12.17\%}$ & $\textbf{13.45}$ & $\textbf{23.17}$ & $\textbf{8.85\%}$ \\
\bottomrule
\end{tabular}
}
\end{table}

\newpage

\section{Extended Interpretability Analysis}
\label{app:interpretability}

This section provides deeper insights into AdaST's learned representations and adaptive gating behavior, complementing the coupling structure analysis and case study in the main paper.

\subsection{Correlation Measure Analysis}

\begin{figure}[t]
    \centering
    \includegraphics[width=\linewidth]{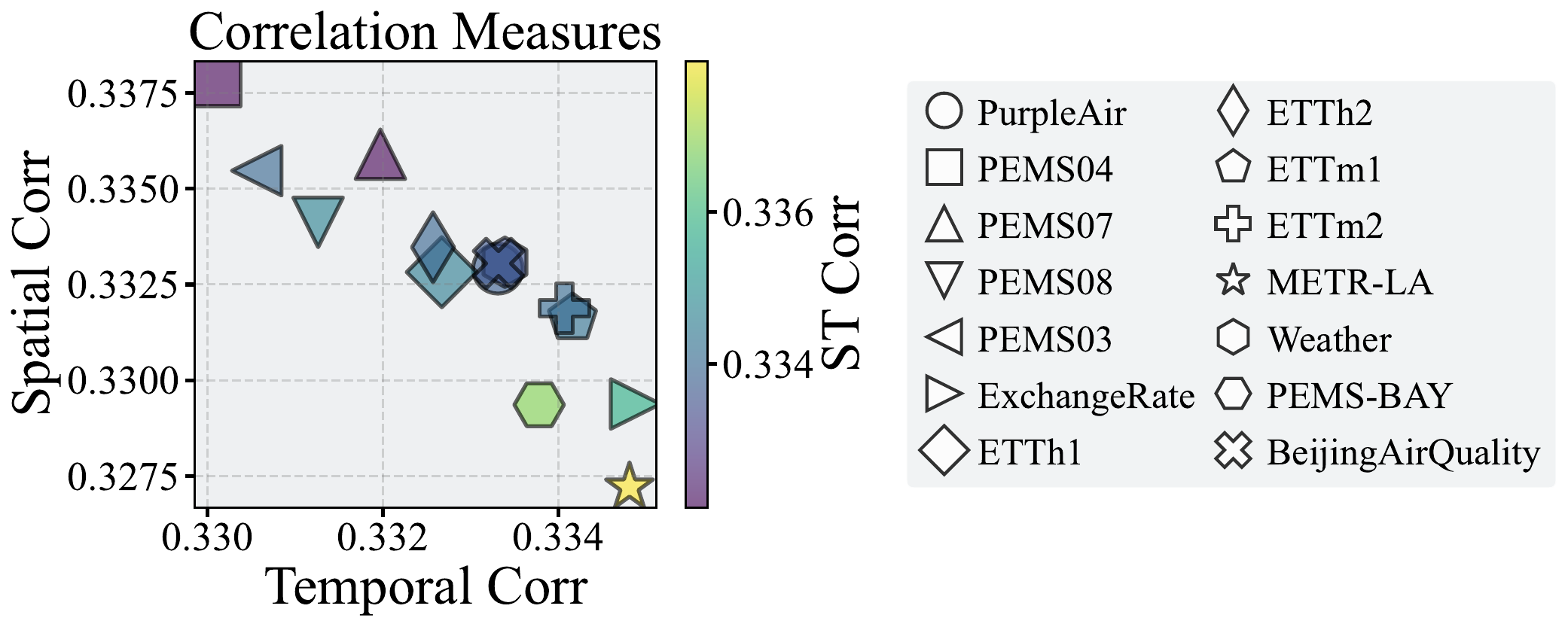}
    \caption{Normalized correlation measures across datasets, further validating the adaptive recomposition mechanism. Strong correspondence with gate scores in Figure~\ref{fig:coupling_structure} confirms that correlation measures reliably reflect each component's information content.}
    \label{fig:corr_scatter}
\end{figure}

We investigate how the normalized correlation measures $\mathcal{C}_t(\mathbf{H}^{(t)})$, $\mathcal{C}_s(\mathbf{H}^{(s)})$, and $\mathcal{C}_{st}(\mathbf{H}^{(st)})$ vary across datasets, providing further insight into the relationship between learned representations and adaptive recomposition. Figure~\ref{fig:corr_scatter} visualizes the average normalized correlation measures across multiple datasets. We observe a strong correspondence between correlation measures and gate scores (Figure~\ref{fig:coupling_structure} in the main paper): datasets with high spatial correlation measures (e.g., PEMS) also exhibit high spatial gate scores, validating our design rationale that both metrics reflect the information content and reliability of each component.

We further identify two systematic relationships among components:
\textbf{(1) Temporal-Spatial Trade-off:} Temporal and spatial correlation measures exhibit a negative relationship --- datasets with stronger temporal correlations tend to show weaker spatial correlations, and vice versa.
\textbf{(2) Temporal-Spatiotemporal Alignment:} Temporal correlation measures show a positive relationship with spatiotemporal correlation measures. This can be attributed to the fact that reduced spatial correlation leads to enhanced temporal and spatiotemporal correlations, with the spatiotemporal component capturing residual joint patterns that complement pure temporal dynamics.

\subsection{Additional Case Studies}

\begin{figure}[t]
    \centering
    \includegraphics[width=\linewidth]{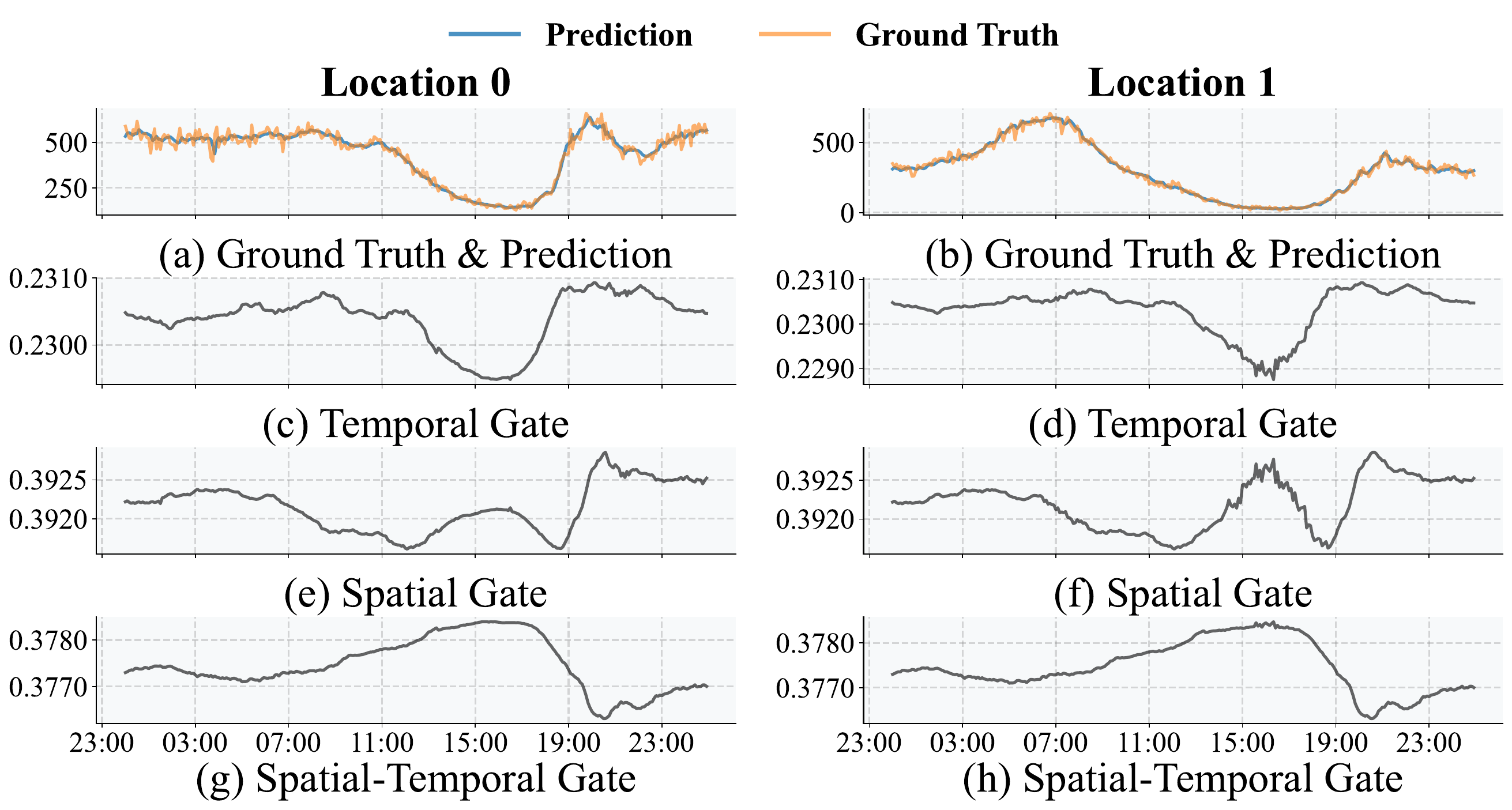}
    \caption{Fine-grained temporal visualization (1-day span) of predictions and gate scores for 2 locations in PEMS07. More accurate predictions correspond to more dynamic gate score trajectories.}
    \label{fig:case_study_s}
\end{figure}

\begin{figure}[t]
    \centering
    \includegraphics[width=\linewidth]{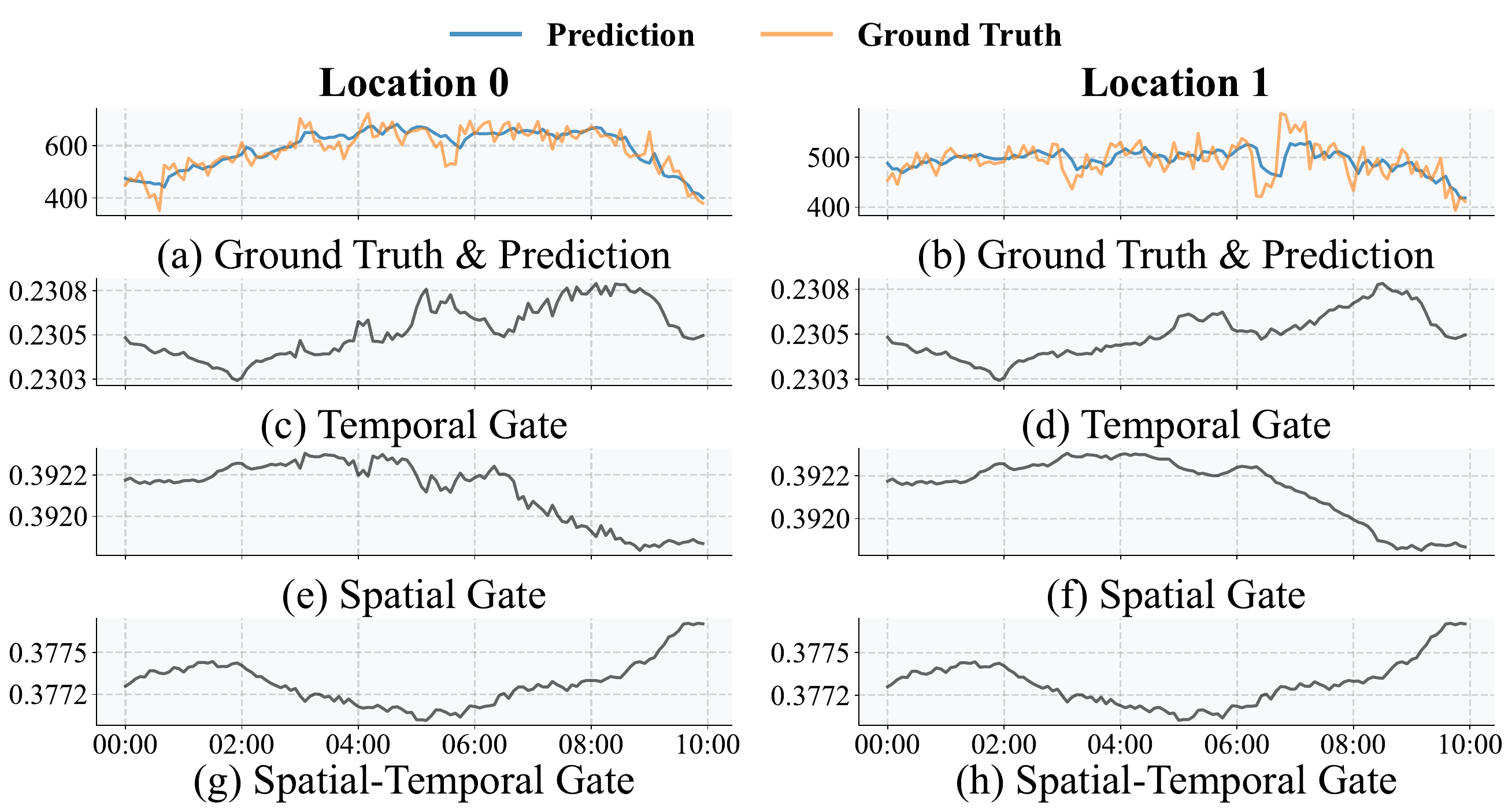}
    \caption{Fine-grained temporal visualization (10-hour span) of predictions and gate scores for 2 locations in PEMS07. Location 0 achieves better accuracy alongside more adaptive gate score dynamics than location 1.}
    \label{fig:case_study_ss}
\end{figure}

\noindent\textbf{Temporal Dynamics of Gate Scores.}
Figures~\ref{fig:case_study_s} and~\ref{fig:case_study_ss} visualize prediction results alongside learned gate scores across fine-grained time periods for two representative locations in PEMS07. Beyond confirming that gate scores vary across time periods (consistent with Figure~\ref{fig:case_study} in the main paper), we observe an interesting phenomenon: more accurate predictions correlate with more dynamic, temporally-varying gate scores. This is particularly evident in Figure~\ref{fig:case_study_ss}, where location $0$ achieves better prediction accuracy than location $1$, accompanied by more turbulent gate score trajectories. We hypothesize that this reflects the model's ability to capture fine-grained coupling dynamics: accurate predictions require adaptive, context-sensitive weighting that responds to local temporal variations in coupling structure, whereas smoother gate scores may indicate insufficient adaptation to changing conditions.

\begin{figure}[t]
    \centering
    \includegraphics[width=\linewidth]{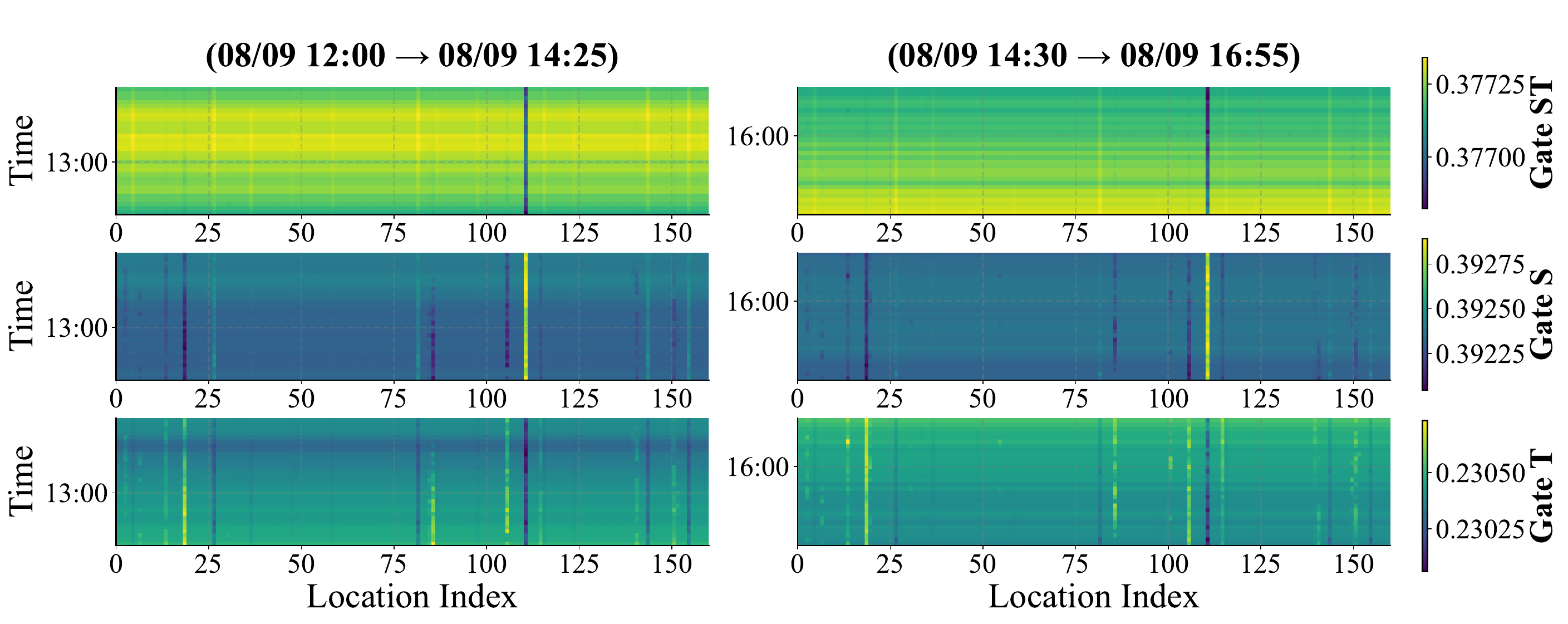}
    \caption{Heatmap of gate scores across 160 spatial locations, showing a globally stable coupling structure with local variations that motivate the use of spatial heterogeneity experts.}
    \label{fig:case_study_spatial}
\end{figure}

\noindent\textbf{Spatial Distribution of Gate Scores.}
Figure~\ref{fig:case_study_spatial} visualizes the learned gate scores across $160$ spatial locations for a fixed time period. We observe two complementary patterns that validate our heterogeneity-aware design: (1) Most locations exhibit similar gate score distributions, indicating a stable dataset-level coupling structure. (2) Despite this overall similarity, certain locations display distinctly different gate score patterns, underscoring the necessity of incorporating spatial heterogeneity experts ($\mathbf{E}_n$) to capture location-specific coupling characteristics.

\newpage
\section{Limitation and Broader Impact}
\label{app:limit}

\subsection{Limitation}
While AdaST demonstrates strong and consistent performance across diverse benchmarks, two limitations remain worth noting. First, the decompose-recompose paradigm introduces additional parameters relative to single-branch architectures, which may increase memory consumption on datasets with very large numbers of nodes. Second, our current evaluation focuses on short-term forecasting with fixed input and output lengths of 12 time steps; extending AdaST to long-term forecasting settings remains an open direction for future work.

\subsection{Broader Impact}
The primary positive impact of this work is enabling more reliable and transparent forecasting in applications such as traffic, air quality, and climate, which can support better planning and resource allocation.
As with many forecasting models, potential risks include misuse for overly confident decision-making, distribution shift leading to degraded performance in deployment, and the amplification of biases or measurement errors present in sensor data. We recommend careful validation under domain-specific conditions, monitoring for performance drift, and transparency about model uncertainty when used in real-world decision pipelines.

\newpage

\input{checklist.tex}

\end{document}

%% file: checklist.tex
\section*{NeurIPS Paper Checklist}

\begin{enumerate}

\item {\bf Claims}
    \item[] Question: Do the main claims made in the abstract and introduction accurately reflect the paper's contributions and scope?
    \item[] Answer: \answerYes{}
    \item[] Justification: The abstract and introduction clearly state the three key contributions of AdaST: (1) identifying coupling structure mismatch as a fundamental challenge in ST forecasting, (2) proposing the decompose-recompose framework with heterogeneity-aware experts and correlation-informed recomposition, and (3) demonstrating state-of-the-art performance across diverse benchmarks. All claims are empirically supported by preliminary experiments in Section~\ref{sec:prelim} and comprehensive evaluations in Section~4.

\item {\bf Limitations}
    \item[] Question: Does the paper discuss the limitations of the work performed by the authors?
    \item[] Answer: \answerYes{}
    \item[] Justification: Limitations are discussed in Appendix~F.1, noting that the decompose-recompose paradigm introduces additional parameters compared to single-branch architectures, and that the current evaluation is restricted to short-term forecasting with fixed horizon length of 12 steps.

\item {\bf Theory assumptions and proofs}
    \item[] Question: For each theoretical result, does the paper provide the full set of assumptions and a complete (and correct) proof?
    \item[] Answer: \answerNA{}
    \item[] Justification: This paper does not include formal theoretical results or proofs. The coupling structure formulations in Section~2.2 are descriptive definitions used to motivate the framework design rather than theorems requiring proof.

\item {\bf Experimental result reproducibility}
    \item[] Question: Does the paper fully disclose all the information needed to reproduce the main experimental results of the paper to the extent that it affects the main claims and/or conclusions of the paper (regardless of whether the code and data are provided or not)?
    \item[] Answer: \answerYes{}
    \item[] Justification: Full implementation details are provided in Section~4.1 and Appendix~\ref{app:impl}, including model architecture, all hyperparameters, optimizer settings, hardware specifications, data splits, preprocessing procedures, and evaluation metrics. An anonymized code repository is available at \url{https://anonymous.4open.science/r/AdaST-70BC}.

\item {\bf Open access to data and code}
    \item[] Question: Does the paper provide open access to the data and code, with sufficient instructions to faithfully reproduce the main experimental results, as described in supplemental material?
    \item[] Answer: \answerYes{}
    \item[] Justification: An anonymized code repository is available at \url{https://anonymous.4open.science/r/AdaST-70BC}. All datasets used (PEMS04/07/08, PurpleAir, METR-LA, ETTh1, ExchangeRate) are publicly available benchmarks, with preprocessing procedures described in Section~4.1 and Appendix~\ref{app:impl}.

\item {\bf Experimental setting/details}
    \item[] Question: Does the paper specify all the training and test details (e.g., data splits, hyperparameters, how they were chosen, type of optimizer) necessary to understand the results?
    \item[] Answer: \answerYes{}
    \item[] Justification: Section~4.1 provides dataset statistics, data splits (60/20/20), Z-score normalization, optimizer settings (Adam, lr=0.001, exponential decay, batch size=16), and all key hyperparameters ($D_H$=256, $L$=3, $\alpha$=0.1). Appendix~\ref{app:impl} further details hyperparameter selection via grid search.

\item {\bf Experiment statistical significance}
    \item[] Question: Does the paper report error bars suitably and correctly defined or other appropriate information about the statistical significance of the experiments?
    \item[] Answer: \answerNo{}
    \item[] Justification: Error bars are not reported due to the high computational cost of running multiple seeds across 16 baselines on 4 datasets. This is consistent with standard practice in the spatial-temporal forecasting community~\cite{shao2024exploring}, and results follow the same single-run evaluation protocol as all compared baselines.

\item {\bf Experiments compute resources}
    \item[] Question: For each experiment, does the paper provide sufficient information on the computer resources (type of compute workers, memory, time of execution) needed to reproduce the experiments?
    \item[] Answer: \answerYes{}
    \item[] Justification: Section~4.1 states that all experiments are conducted on an NVIDIA A100 80GB GPU. Per-epoch training times for all four main benchmarks are reported in Table~\ref{tab:efficiency} in Appendix~\ref{app:impl}.

\item {\bf Code of ethics}
    \item[] Question: Does the research conducted in the paper conform, in every respect, with the NeurIPS Code of Ethics \url{https://neurips.cc/public/EthicsGuidelines}?
    \item[] Answer: \answerYes{}
    \item[] Justification: This work uses only publicly available sensor datasets for forecasting tasks and involves no human subjects, sensitive personal data, or applications with direct harm potential. The research fully conforms with the NeurIPS Code of Ethics.

\item {\bf Broader impacts}
    \item[] Question: Does the paper discuss both potential positive societal impacts and negative societal impacts of the work performed?
    \item[] Answer: \answerYes{}
    \item[] Justification: Broader impacts are discussed in Appendix~F.2. Positive impacts include improved forecasting reliability for traffic management, air quality monitoring, and climate applications. Potential risks include overconfident decision-making, performance degradation under distribution shift, and amplification of sensor biases, along with recommended mitigation strategies.

\item {\bf Safeguards}
    \item[] Question: Does the paper describe safeguards that have been put in place for responsible release of data or models that have a high risk for misuse (e.g., pre-trained language models, image generators, or scraped datasets)?
    \item[] Answer: \answerNA{}
    \item[] Justification: This paper proposes a general spatial-temporal forecasting framework trained on public sensor datasets. It does not release pre-trained generative models or scraped data that carry significant misuse risk.

\item {\bf Licenses for existing assets}
    \item[] Question: Are the creators or original owners of assets (e.g., code, data, models), used in the paper, properly credited and are the license and terms of use explicitly mentioned and properly respected?
    \item[] Answer: \answerYes{}
    \item[] Justification: All datasets and baseline methods are properly cited in Section~4.1 and the references. Baselines are evaluated using their official implementations with recommended hyperparameters, as stated in Section~4.1.

\item {\bf New assets}
    \item[] Question: Are new assets introduced in the paper well documented and is the documentation provided alongside the assets?
    \item[] Answer: \answerYes{}
    \item[] Justification: The AdaST codebase is released at \url{https://anonymous.4open.science/r/AdaST-70BC} with documentation covering model architecture, training procedures, and scripts to reproduce all main experimental results reported in the paper.

\item {\bf Crowdsourcing and research with human subjects}
    \item[] Question: For crowdsourcing experiments and research with human subjects, does the paper include the full text of instructions given to participants and screenshots, if applicable, as well as details about compensation (if any)?
    \item[] Answer: \answerNA{}
    \item[] Justification: This paper does not involve crowdsourcing or research with human subjects. All experiments are conducted on publicly available sensor and traffic datasets.

\item {\bf Institutional review board (IRB) approvals or equivalent for research with human subjects}
    \item[] Question: Does the paper describe potential risks incurred by study participants, whether such risks were disclosed to the subjects, and whether Institutional Review Board (IRB) approvals (or an equivalent approval/review based on the requirements of your country or institution) were obtained?
    \item[] Answer: \answerNA{}
    \item[] Justification: This paper does not involve human subjects and therefore requires no IRB approval or equivalent review.

\item {\bf Declaration of LLM usage}
    \item[] Question: Does the paper describe the usage of LLMs if it is an important, original, or non-standard component of the core methods in this research?
    \item[] Answer: \answerNA{}
    \item[] Justification: LLMs are not used as any part of the core methodology. Any incidental use of LLMs was limited to writing assistance only and does not affect the scientific contributions, experimental design, or reported results.

\end{enumerate}